\documentclass[conference]{IEEEtran}
\IEEEoverridecommandlockouts
\usepackage{cite}
\usepackage{amsmath,amssymb,amsfonts}
\usepackage{algorithmic}
\usepackage{graphicx}
\usepackage{textcomp}

\usepackage{latexsym}
\usepackage{verbatim}
\usepackage{url}

\usepackage{xcolor}
\usepackage{bm}
\usepackage{amsmath,amsthm}

\begin{document}

\title{Distinguishing between Normal and Cancer Cells Using Autoencoder Node Saliency*\\
\thanks{This work has been supported in part by the Joint
Design of Advanced Computing Solutions for Cancer
(JDACS4C) program established by the U.S. Department
of Energy (DOE) and the National Cancer Institute (NCI)
of the National Institutes of Health.}
}

\author{\IEEEauthorblockN{Ya Ju Fan\IEEEauthorrefmark{1},
Jonathan E. Allen\IEEEauthorrefmark{2}, 
Sam Ade Jacobs\IEEEauthorrefmark{1}, and
Brian C. Van Essen\IEEEauthorrefmark{1}}
\IEEEauthorblockA{\IEEEauthorrefmark{1}Center for Applied Scientific Computing, Lawrence Livermore National Laboratory, Livermore, CA, USA \\ Email: fan4@llnl.gov}
\IEEEauthorblockA{\IEEEauthorrefmark{2}Computation Directorate, Lawrence Livermore National Laboratory, Livermore, CA, USA}}

\maketitle

\begin{abstract}
Gene expression profiles have been widely used to characterize patterns of cellular responses to diseases. As data becomes available, scalable learning toolkits become essential to processing large datasets using deep learning models to model complex biological processes. We present an autoencoder to capture nonlinear relationships recovered from gene expression profiles. The autoencoder is a nonlinear dimension reduction technique using an artificial neural network, which learns hidden representations of unlabeled data. We train the autoencoder on a large collection of tumor samples from the National Cancer Institute Genomic Data Commons, and obtain a generalized and unsupervised latent representation. We leverage a HPC-focused deep learning toolkit, Livermore Big Artificial Neural Network (LBANN) to efficiently parallelize the training algorithm, reducing computation times from several hours to a few minutes. With the trained autoencoder, we generate latent representations of a small dataset, containing pairs of normal and cancer cells of various tumor types. A novel measure called autoencoder node saliency (ANS) is introduced to identify the hidden nodes that best differentiate various pairs of cells. We compare our findings of the best classifying nodes with principal component analysis and the visualization of t-distributed stochastic neighbor embedding. We demonstrate that the autoencoder effectively extracts distinct gene features for multiple learning tasks in the dataset.

\end{abstract}

%============================================
%
\section{Background and motivation}
%
%============================================	

Cancer is a group of genetic diseases, characterized by the development of abnormal cells that have the ability to infiltrate and destroy normal body tissues \cite{NCI2015}. Cancer genome sequencing has fundamentally improved our understanding of mutations in cancer cells. The discoveries through cancer genome sequencing show that mutational processes vary among tumors and cancer types \cite{Martincorena2015, Vogelstein2013}. Identifying the genomic alterations that arise in cancer can help doctors select treatments based on their distinct molecular abnormalities. Interesting findings in gene expression profiles provide implications of how normal cells evolve into cancer cells.  Zhang et al. \cite{Zhang1997} analyzed gene expression patterns in gastrointestinal tumors, and Welsh et al. \cite{Welsh2001} identifies candidate molecular markers of epithelial ovarian cancer. Most studies employ direct comparisons between normal and cancer cells using statistical measurements.

Large-scale gene expression profiles can consist of thousands of features. To reduce the cost of profiling and satisfy a reasonable number of features for limited sample sizes \cite{Raudys1991, Hua2005, Dobbin2008}, a subset of genes are frequently pre-selected based on high variability \cite{Peck2006, Skov2012}. However, advances in sequencing now allow for more profiling of a larger set of genes in non-clinical diagnostic settings. Research has shown that deep learning is able to handle a larger number of features and perform better than the linear methods used for selecting landmark genes \cite{Chen2016}. It implies that deep learning could capture complex nonlinear relationships between expressions of genes missed by linear methods. Although kernel machines can represent useful nonlinear patterns \cite{Ye2013}, they do not scale well to the growing data size of expression profiles. Thus, deep learning enjoying both representability and scalability is ideal for large-scale non-linear gene expression feature extraction \cite{Chen2016}.

A massive proportion of cell lines used in biomedical research is mislabeled \cite{Allen2016, ATC2010}, causing potentially erroneous findings. Supervised learning uses class labels, i.e. the responses of the observations, to guide the training algorithm and can be successful in data classification. However, its performance is sensitive to the quality of the labels. 

Motivated by these results, we demonstrate the potential of applying deep learning methods on large-scale gene expression profiles. We focus on the unsupervised learning method, called the autoencoder, which is an artificial neural network that transforms data into a meaningful latent space. Since no tumor type labels are considered when training the autoencoder, the algorithm avoids being misled by false labeling. It has been applied on genome-wide assays of cancer for knowledge extraction using their unsupervised nature. The denoising autoencoders generate latent representations to classify breast cancer cells from normal control cells \cite{Tan2015, Danaee2017}. Variational autoencoders are built on the genomic data \cite{Way2018} where explanatory features are obtained by subtracting a series of mean values of latent representations to obtain insights linking specific features to biological pathways. These models are built on selected gene sets, ranging from 2,520 to 15,000 genes. Existing autoencoder methods do not give a rigorous metric for rank ordering hidden network nodes to explore and operate on small training sets.

%The feature extraction is based on known properties. It lacks a tool to guide ...

%We apply an autoencoder to extract distinct expression features in the latent space. 
The novelty of this work is in the following. First, we apply transfer learning \cite{Baxter1998} to overcome the problems of high dimension, low sample size data \cite{Bellman1957, Dobbin2008}. We train the autoencoder on a large collection of tumor samples to obtain a generalized latent representation. We include all 60,483 measured transcripts, which requires a large amount of computing resources. With its optimal parameters, we compute latent representations of a small dataset containing pairs of normal and cancer cells. Second, we use the latent space to characterize complex gene expression activations existing in different tumor types. This is done by applying the autoencoder node saliency (ANS) \cite{Fan2019} to rank the hidden projections according to their ability to classify normal and cancer cells of different tissue types. Furthermore, we leverage a HPC-focused deep learning toolkit, Livermore Big Artificial Neural Network (LBANN) to efficiently parallelize the training algorithm, reducing computation times from several hours to a few minutes.

%we include all available genes in our data when training the autoencoder.

%===========================================
%
\section{Data Description}
%
%===========================================
The National Cancer Institute Genomic Data Commons (GDC) data portal provides unified genomic data from patients with cancer \cite{Grossman2016}. We obtained two datasets for our analysis,  a collection of pairs of normal and cancer cells and a large collection of GDC tumor samples. Both datasets contain cell lines measured by the same 60,483 transcripts.  The first dataset contains pairs of normal and cancer cells from 23 cancer types. Since some cancer types have very small number of samples, we select tumor types that have larger than 40 pairs of normal and cancer cells, making a total of 533 cell pairs across nine cancer types for our experiments. Careful attention was given to exclude samples in the test dataset from samples in the training dataset, and in particular, all cancer-normal pairs were excluded from training. A detailed description of these tumor types and their numbers of samples are shown in Table~\ref{tb:dataDescription}. 

Since the first dataset has a relatively larger number of features (60,483) to the number of samples (1,066), we need a sufficient number of training samples in order to obtain a good performing neural network \cite{Hua2005}. Therefore, the second dataset is designated for training the autoencoder for obtaining a more generalized latent representation. It consists of 11,574 tumors from 28 cancer types.  
% JEA I don't understand the purpose of the last sentence, are you saying that 11,574 samples is sufficient?
%JEA there should be a description of where the second dataset comes from. Is this a subset of GDC ? The main thing is to make clear that it is (hopefully) non-overlapping with the training set.

\begin{figure*}[htbp]
\begin{center}
\includegraphics[width = 0.9\linewidth, trim=0cm 0cm 0cm 0cm, clip=true]{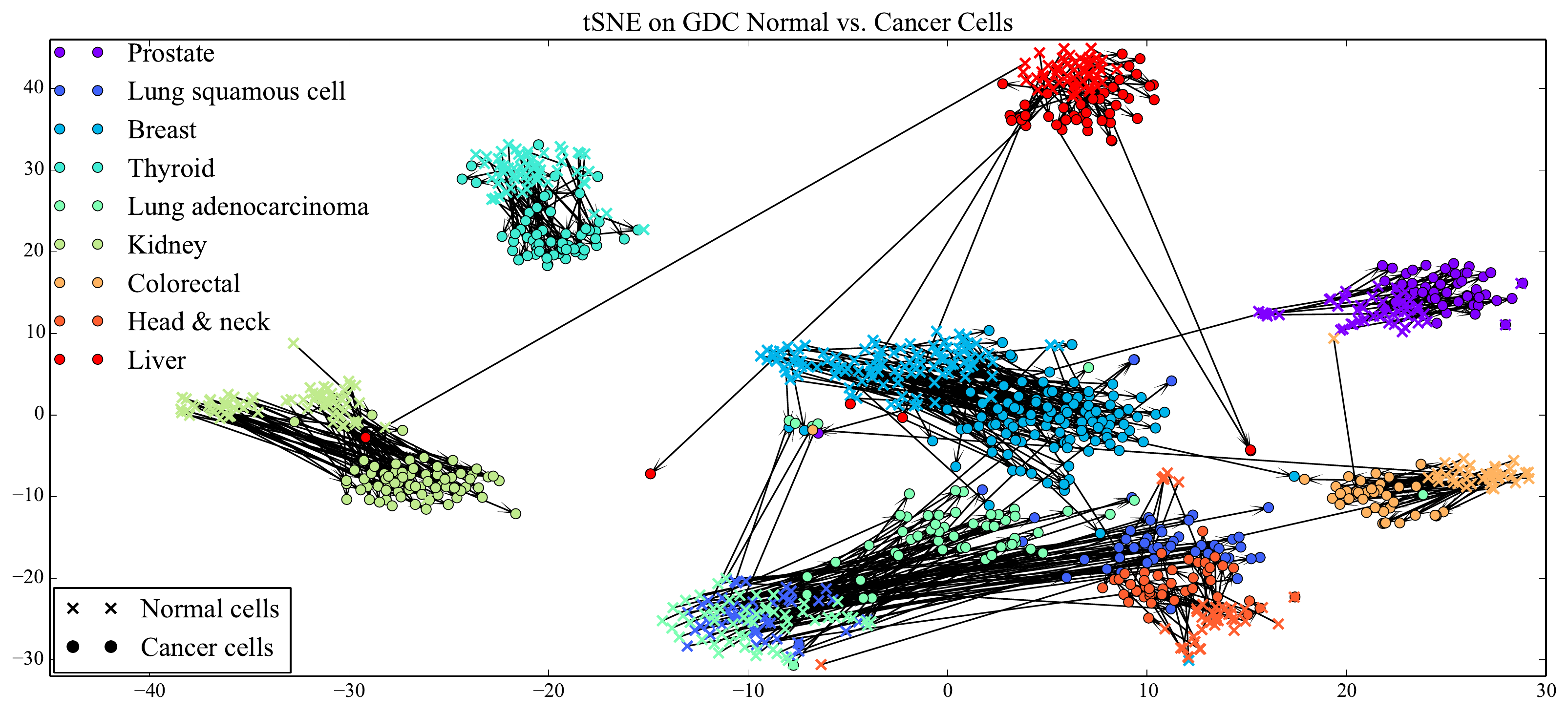}
\vspace{-0.2in}
\caption{Visualization of the normal and cancer cell pairs using two dimensional tSNE.}
\label{fg:tSNE}
\end{center}
\end{figure*}

In Figure~\ref{fg:tSNE}, we visualize the distribution of the normal and cancer cell pairs using t-distributed stochastic neighbor embedding (t-SNE). The t-SNE \cite{Maaten2008} is a nonlinear dimension reduction technique that projects the high-dimensional data in two-dimensional space. It locates similar high-dimensional data points next to each other in the low dimension, at the same time puts dissimilar high-dimensional data points away from each other. It largely avoids overlapping data points for visualization. Although the euclidean distance is used as the base of its similarity metric in the original space, the distances are not preserved. Therefore, the projection in the lower dimension is only for understanding the similarity and dissimilarity of the data clusters. 

Since tumor types are not considered when generating the t-SNE plot, in Figure~\ref{fg:tSNE}, we see that each tumor type naturally forms a cluster, and within the same tumor types the normal cells are gathered in a cluster and the cancer cells are gathered in another. Each pair of the normal and cancer cells is connected with an arrow. The arrows identify data points far away from their tumor type clusters where data points could have been mislabeled. For example, there is a small cluster between the breast cluster and the lung cluster, which contains mixed cancer cells that are difficult to classify.

\begin{table}[htbp]
\caption{Description of the cancer and normal cell pairs and their corresponding number of samples. We consider the tumor types that have more than 40 samples.}
\begin{center}
\begin{tabular}{|c|c|c|r|}
\hline
\textbf{Cancer Cell} & \textbf{Normal Cell} & \textbf{Count} \\
\hline
Breast Invasive Carcinoma
	&Breast &112\\
Prostate Adenocarcinoma 
	&Prostate & 51\\
Lung Squamous Cell Carcinoma
	&Lung &49\\
Lung Adenocarcinoma
	&Lung & 57\\
Thyroid Carcinoma
	&Thyroid  & 58\\
Kidney Renal Clear Cell Carcinoma
	&Kidney  & 72\\
Colon Adenocarcinoma
	&Colorectal  & 41\\
Head and Neck Squamous Cell Carcinoma
	&Head and Neck  & 43\\
Liver Hepatocellular Carcinoma
	&Liver & 50\\
\hline
\end{tabular}
\label{tb:dataDescription}
\end{center}
\end{table}

%We randomly sample 80 percent of the tumors in the large GDC dataset and train the autoencoder on the sampled subset. 

%===============================

%===============================
%
\section{Autoencoder Node Saliency}
%
%===============================
%---------------------
%
\subsection{Notation}
%
%---------------------
In this paper, matrices are denoted by bold uppercase letters, vectors by bold lowercase letters and scalars by letters not in bold. Let $\bm{A}$ be a matrix whose $i$-th row vector is $\bm{a}_i$, and $\bm{b}$ be a vector whose $s$-th element is $b_s$. The element at the $i$-th row and the $j$-th column of $\bm{A}$ is $a_{i,j}$. Let $\bm{1}$ be a vector of all ones with a suitable dimension. The cardinality of a set $T$ is $|T|$. We denote a dataset as $\bm{X} \in [0,1]^{n\times d}$ that contains $n$ data points $\bm{x}_i \in [0,1]^d$ of $d$-dimensional variables for $i=1,\dots,n$.

%----------------------------------
%
\subsection{Autoencoder}
\label{sec:ae}
%
%----------------------------------
Autoencoder builds a lower-dimensional representation of the data through a pair of maps $\bm{X} \overset{f}{\rightarrow} \bm{A} \overset{g}{\rightarrow} \bm{X}$. The first map is the encoder $\bm{A} = f(\bm{X})$ and the second is the decoder $\bm{X'} = g(\bm{A})$.
%JEA is the activation values really always binary here? I don't understand the emphasis on [0,1] and not a \REAL value.
Using $m$ hidden nodes in the neural network, the encoder performs dimension reduction on the input $\bm{X}$ and transforms the data of dimension $d$ to a reduced dimension $m$ where $m<d$; the decoder performs a reconstruction, mapping the data from the reduced dimension $m$ back to the original dimension $d$. This is done so that the reconstruction error between $\bm{X}$ and $g(f(\bm{X}))$ is small. Usually autoencoders are restricted so that they do not simply learn the input set perfectly, but prioritize which aspects of the input should be kept. Hence, autoencoders often learn useful properties of the data \cite{Goodfellow2016}.

The latent representation of the dataset $\bm{X}$ is defined as
\begin{equation}
\bm{A} = f(\bm{X}) = \sigma(\bm{W}\bm{X}^{\mathsf{T}} + \bm{b}),
\label{eq:activation}
\end{equation}
\noindent where $\bm{W}\in\mathbb{R}^{m\times d}$ and $\bm{b}\in\mathbb{R}^m$ are weight vector and bias in the linear transform; and $\sigma$ is the activation function in the nonlinear transform. Most activation functions in neural networks try to capture the rate of action potential, whose simplest form is a binary function. We use the sigmoid function, $\sigma(z) = (1+e^{-z})^{-1}$, to handle such design where the action frequency increases quickly at first, but gradually approaches an asymptote at 100 percent action. 

The decoder then maps the activation $\bm{A}$ to the reconstruction $\bm{X'}$ to the same dimensional space of $\bm{X}$,
\[
\bm{X'}=g(\bm{A})=\sigma(\bm{W^{\mathsf{T}}A}+\bm{b'})^{\mathsf{T}},
\]
\noindent where $\bm{b'}\in\mathbb{R}^d$ is the bias term in the decoder. We train the autoencoder on the GDC dataset by finding optimal solutions for $\bm{W}^*$, $\bm{b}^*$ and $\bm{b'}^*$ that minimize the mean squared error, the difference between $\bm{X}$ and $\bm{X'}$. %An autoencoder is driven to capture the most salient features of the input data by restricting the algorithm from simply learning the dataset $g(f(\bm{X})) = \bm{X}$.
%JEA HOW IS THE ALGORITHM RESTRICTED? IT's NOT CLEAR FROM THIS, ARE YOU TALKING ABOJT A DENOISING, DROPOUT OR OTHER REGULARIZATION?
%In practice, the dataset is divided in mini-batches for using the stochastic gradient descent algorithm to find the optimal solution that minimizes the reconstruction error.

After training an autoencoder on the GDC dataset $\bm{X}$, we are interested in identifying features in the autoencoder that can be used to distinguish between normal and cancer cells . Given a hidden node $s$ (for $s=1,\dots,m$), we transform the normal and cancer cell pairs $\bm{\hat{X}}$ in the GDC features to their latent representations (also called activation values): 
\begin{equation}
\bm{a}_s=\sigma(\bm{w}_s^*\bm{\hat{X}}^{\mathsf{T}}+b_s^*\cdot\bm{1}).
\label{eq:activation}
\end{equation}

Autoencoder node saliency (ANS) \cite{Fan2019} is based on the histograms of the activation values. The vector $\bm{a}_s$ at node $s$ can be described using a one-dimensional histogram.  The activation values are in the range of $(0,1)$, due to our selection on the activation function, the sigmoid, that restricts the range of projection. A histogram contains a set of $k$ bin ranges, $B = \{[0,\frac{1}{k}),[\frac{1}{k},\frac{2}{k}),\dots,[\frac{k-1}{k},1]\}$, where $[a,b)$ indicates values $\geq a$ and $<b$. Simply, the $r$-th bin range in the set $B$ is $B_r = [\frac{r-1}{k},\frac{r}{k})$ for $r=1,\dots,k$. The ANS method contains two parts. The first part is the unsupervised node saliency that measures the ``interestingness'' of the histograms, using normalized entropy difference (NED). The second part is the supervised node saliency (SNS) that incorporates the distribution of the two class labels in the histograms.

%----------------------------------------
%
\subsection{Unsupervised node saliency}
\label{sec:NED}
%
%----------------------------------------

Given the constructed histogram, we first compute the entropy of the latent representations at the hidden node $s$ for $s=1,\dots,m$, defined by: 
\[
E(\bm{a}_s) = -\sum_{r}p(B_r,\bm{a}_s)\log_2 p(B_r,\bm{a}_s);
\]
\noindent where $|\bm{a}_s| = \hat{n}$, the number of data points encoded, and
\begin{equation}
p(B_r,\bm{a}_s)\equiv \frac{\left|\{i\mid a_{s,i}\in B_r \}\right|}{|\bm{a}_s|}, 
\label{eq:pBr}
\end{equation}
is the probability of the activation values $\bm{a}_s$ occurring in the $r$-th bin range of the histogram for $r=1,\dots,k$. To compare different histograms, the unsupervised node saliency utilizes the normalized entropy difference (NED) %\cite{Rognvaldsson2017}, 
defined as
\begin{equation}
\label{eq:NED}
\begin{split}
\text{NED}(\bm{a}_s) & = \displaystyle\frac{\log_2 \hat{k} - E(\bm{a}_s)}{\log_2 \hat{k}} \\
                & = 1 + \frac{1}{\log_2 \hat{k}}\sum_{r}p(B_r,\bm{a}_s)\log_2 p(B_r,\bm{a}_s),
\end{split}
\end{equation}

\noindent where $\hat{k}$ is the number of occupied bins. Note that $\log_2\hat{k}$ is the maximum entropy of $\bm{a}_s$ ($E(\bm{a}_s)$) when $p(B_r,\bm{a}_s)=1/\hat{k}$ for all occupied bins. 

After training the autoencoder without considering class labels, we examine whether the features constructed by the autoencoder exhibit properties related to known class labels, which are the normal cells labeled as 0 and the cancer cells labeled as 1. We define $p_c(B_r,\bm{a}_s)$ as the probability that the activation values from one of the two classes, $c\in\{0,1\}$ occur in the bin range $B_r$, which is
\[
p_c(B_r,\bm{a}_s) \equiv \frac{\left|\{i\mid a_{s,i}\in B_r \quad\text{and}\quad y_i=c\}\right|}{n_c},
\]
\noindent where $n_c$ is the number of data points in class $c$. Using (\ref{eq:NED}) for each of the classes, we obtain the supervised NED defined as:
\begin{equation}
\label{eq:NEDclass}
\text{NED}_c(\bm{a}_s)=1 + \frac{1}{\log_2 \hat{k}}\sum_{r}p_c(B_r,\bm{a}_s)\log_2 p_c(B_r,\bm{a}_s).
\end{equation}
The autoencoder node saliency evaluates three values: NED  defined in (\ref{eq:NED}) on the data of the two classes, 0 and 1, combined; as well as $\text{NED}_0$ and $\text{NED}_1$ defined in (\ref{eq:NEDclass}), each on the data from one of the two classes. 

Note that the range of NED is between zero and one. When NED equals one, the node's activation value is near constant, occupies only one bin and low information content. A high value of NED (below 1) indicates that most of the activation values are settled in a few bins giving an ``interesting'' profile. A low value of NED corresponds with activation values spread evenly over all the bins, indicating ``uninteresting'' nodes. Node saliency increases with NED, except for the extreme case when NED equals one. Therefore, a good classifying node has activation values from one class take up a few bins in the histogram, with a different set of bins occupied by the other class. Since the activation values from both classes occupy the union of these bins, the data distribution combining both classes is less ``interesting'' than the data distribution of each individual class. A good classifying node has a property that satisfies both NED $<\text{NED}_0$ and NED $<\text{NED}_1$ \cite{Fan2019}.

%-----------------------------------------
%
\subsection{Supervised node saliency}
%
%-----------------------------------------

The NED values reveal the ``interestingness'' of the latent representations at each of the autoencoder hidden nodes. To rank the hidden nodes according to their capability of separating the normal and cancer cells, we apply the supervised node saliency (SNS) \cite{Fan2019} that measures the binomial proportions of the two classes in the histogram. The SNS compares a distribution $\bm{q}$ against a fixed reference distribution $\bm{p}$. It is at its minimal value when $\bm{q}=\bm{p}$. The distribution $\bm{q}$ varies at different hidden nodes. It is constructed using the binomial proportions at each bin range $r$ of activation values at node $s$ defined in (\ref{eq:qr}).
\begin{equation}
\begin{split}
q_r &= \text{prob}\{y_i=1 \mid a_{s,i} \in B_r\} \\
    &\equiv \frac{\left|\{i\mid a_{s,i}\in B_r \quad\text{and}\quad y_i = 1\}\right|}{\left|\{i\mid a_{s,i}\in B_r\}\right|}. 
\end{split}
\label{eq:qr}
\end{equation}
The fixed reference distribution $\bm{p}$ is designed manually using binary distribution defined in (\ref{eq:binary}), where one class totally occupies half of the bins at one end, and the other class falls on the other half of the bins.
\begin{equation}
p_r = 
\begin{cases}
0 & \quad \text{if } r < k/2\\
1 & \quad \text{if } r \geq k/2.
\end{cases}
\label{eq:binary}
\end{equation}

The SNS applies weighted cross entropy (WCE) employing a weight, $p(B_r,\bm{a}_s)$, at each bin $r$.  For notation convenience, we let $p(B_r) = p(B_r,\bm{a}_s)$. The SNS is defined as 
\begin{equation}
\begin{split}
\text{SNS} \equiv \min \Big\{\text{WCE}_0, \text{WCE}_1\Big\}, \text{where}
\end{split}
\label{eq:sns}
\end{equation}
\[
\text{WCE}_1=\sum_{r} p(B_r)\Big[-p_r\log_2 q_r - (1-p_r)\log_2(1-q_r)\Big]\text{and}
\]
\[
\text{WCE}_0=\sum_r p(B_r)\Big[-(1-p_r)\log_2 q_r - p_r\log_2(1-q_r)\Big].
\]
\noindent The SNS gives one value for each of the hidden nodes measuring the similarity of their binomial distributions on the two classes to the binary distribution $\bm{p}$. A smaller value of SNS at a hidden node indicates that its class distribution is closer to $\bm{p}$, where the two classes are perfectly separated. Thus, we rank the hidden nodes in the ascending order of SNS. 

The autoencoder node saliency identifies best classifying hidden nodes and contains two steps. The first is to order the hidden nodes according to the SNS, and the next is to interpret the latent representations based on their NED values.

%====================================
%
\section{Parallelization with LBANN}
\label{sec:lbann}
%====================================
%----------------------------------------------
%
\subsection{LBANN overview}
\label{sec:lbann:overview}
%
%----------------------------------------------
The deep learning algorithm described in Section~\ref{sec:ae} was implemented using the
Livermore Big Artificial Neural Network (LBANN) toolkit. LBANN is an open
source framework being developed at US DOE Lawrence Livermore National
Laboratory for training deep neural networks at scale (that is,
on very large data sets and with large models) with a specific focus
on HPC environments ~\cite{van2015lbann}. As an HPC-centric
toolkit, LBANN is optimized for both strong and weak scaling and it is
designed to exploit multiple parallelism schemes: intra- and
inter-model data parallelism (distributing data samples across
processes) and model parallelism (distributing model parameters across
processes).% (see Figure \ref{fig:lbann-parallelism}).

\begin{comment}
\begin{figure*}[htp]
  \centering
  \includegraphics[width=1\columnwidth, trim=1.2cm 1.2cm 1cm 0cm, clip=true]{plots/scaling-nodeselect-catalyst-surface.pdf}
  \vspace{-0.28in}
  \caption{Intra-model data parallelism, model parallelism, and
    inter-model data (hybrid) parallelism in LBANN.}
  \label{fig:lbann-parallelism}
\end{figure*}
\end{comment}

At the core of LBANN is the Elemental math library
~\cite{poulson2013elemental}. Elemental is MPI-based and it provides
highly optimized data structures and algorithms for distributed linear
algebra. It handles most of the distributed data management and its
distributed matrix multiplication routine does the heavy lifting for
model parallel computation. When GPUs are available, LBANN uses custom
CUDA kernels and optimized cuDNN routines to perform much of the
computation. LBANN's data readers have support for data staging
through node-local NVRAM and can ingest data in various forms.

%-----------------------------------------
%
\subsection{Scaling results}
\label{sec:lbann:results}
%
%-----------------------------------------

Scaling experiments were run on the Surface and Catalyst clusters at Lawrence Livermore National Laboratory~\cite{lchardware}. 
Surface consists of 156 compute nodes, each with a 16-core Intel Sandy Bridge
CPU, 256 GB memory, and two Tesla K40 GPUs. Catalyst consists of 324
nodes, each with two 12-core Intel Xeon E5-2695 v2 CPUs, 120 GB RAM,
800 GB NVRAM, and dual InfiniBand QDR network interfaces.

\begin{figure}[th]
  \centering
  \includegraphics[width=\columnwidth,trim=0.4cm 0.6cm 0cm 0cm, clip=true]{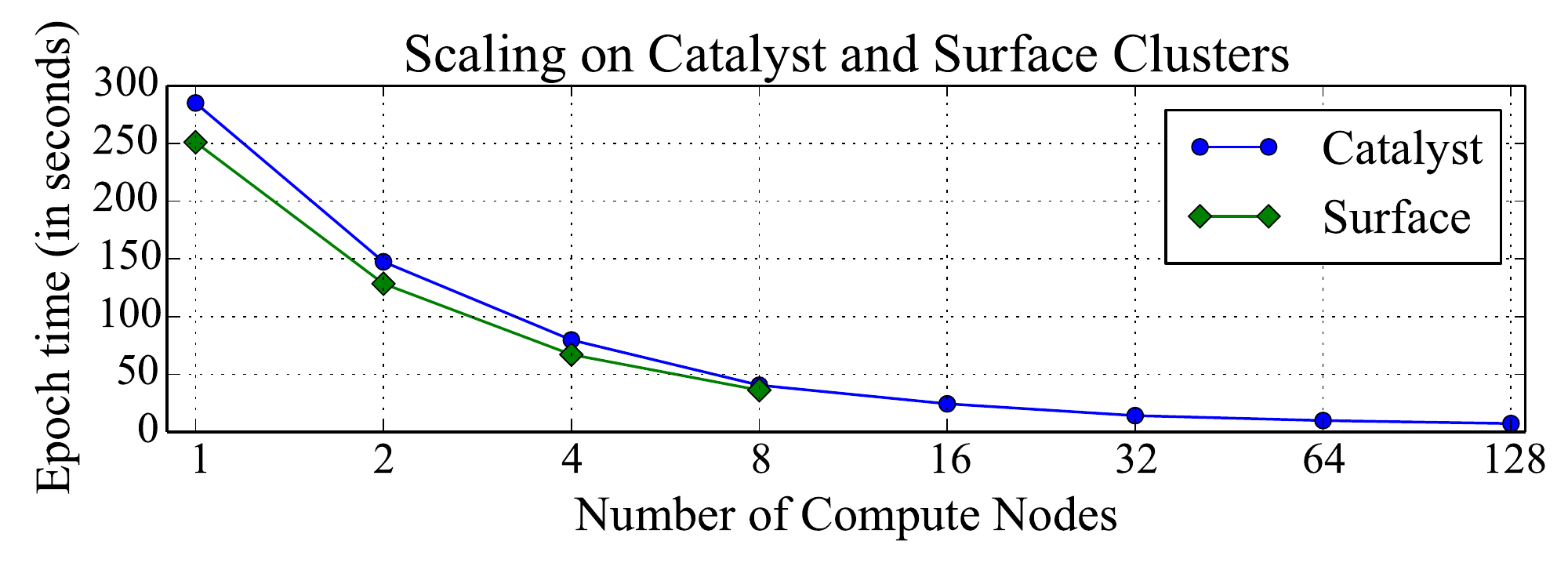}
  \vspace{-0.28in}
  \caption{Strong scaling results on Catalyst (CPU cluster) and Surface (GPU cluster)}
  \label{fig:lbann-scaling}
\end{figure}

Strong scaling experiments were conducted on both Surface and Catalyst HPC
clusters by varying the number of compute nodes with the
problem size fixed. Figure~\ref{fig:lbann-scaling} shows
strong scaling results of training the autoencoder model on up to 8 and 128 computer
nodes for Surface and Catalyst cluster respectively.  We observed reductions in
the running time as the number of compute nodes increased. The modest strong 
scaling reduces the overall computation time to convergence of the autoencoder
model from several hours to a few minutes.   

%==========================================
%
\section{Experimental Results}
%
%==========================================

%----------------------------------------------------
%
\subsection{Training the autoencoder}
%
%----------------------------------------------------

To determine the appropriate parameter settings for the GDC dataset, we perform a full factorial design over all combinations of selected parameter values. Details of the training method can be found in Chapter eight of the deep learning book \cite{Goodfellow2016}. We started at a wide range of random parameter selections and then narrowed down to those that give better performance. The parameters are: hidden layer size of 500 and 1000; batch size of 25, 50, 100 150 and 200; and learning rate of 0.00001, 0.00005, 0.0001, 0.00025, 0.0005, 0.00075 and 0.001. The best epoch sizes are chosen when the cost function stops decreasing and the Pearson correlation of the reconstruction stops increasing. We split the dataset into a training set and a validation set. We use the validation set to evaluate the model during the training process and limit overfitting the model. The best parameter combination was batch size 200 and learning rate 0.00005 at 230 epochs with 1000 hidden nodes. This setting gives a Pearson correlation of 0.8574 on the validation set. 

%----------------------------------------------------
%
\subsection{Principal components analysis}
\begin{figure}[th]
\begin{center}
\includegraphics[width = 1.0 \columnwidth, trim=0cm 0.2cm 0cm 0cm, clip=true]{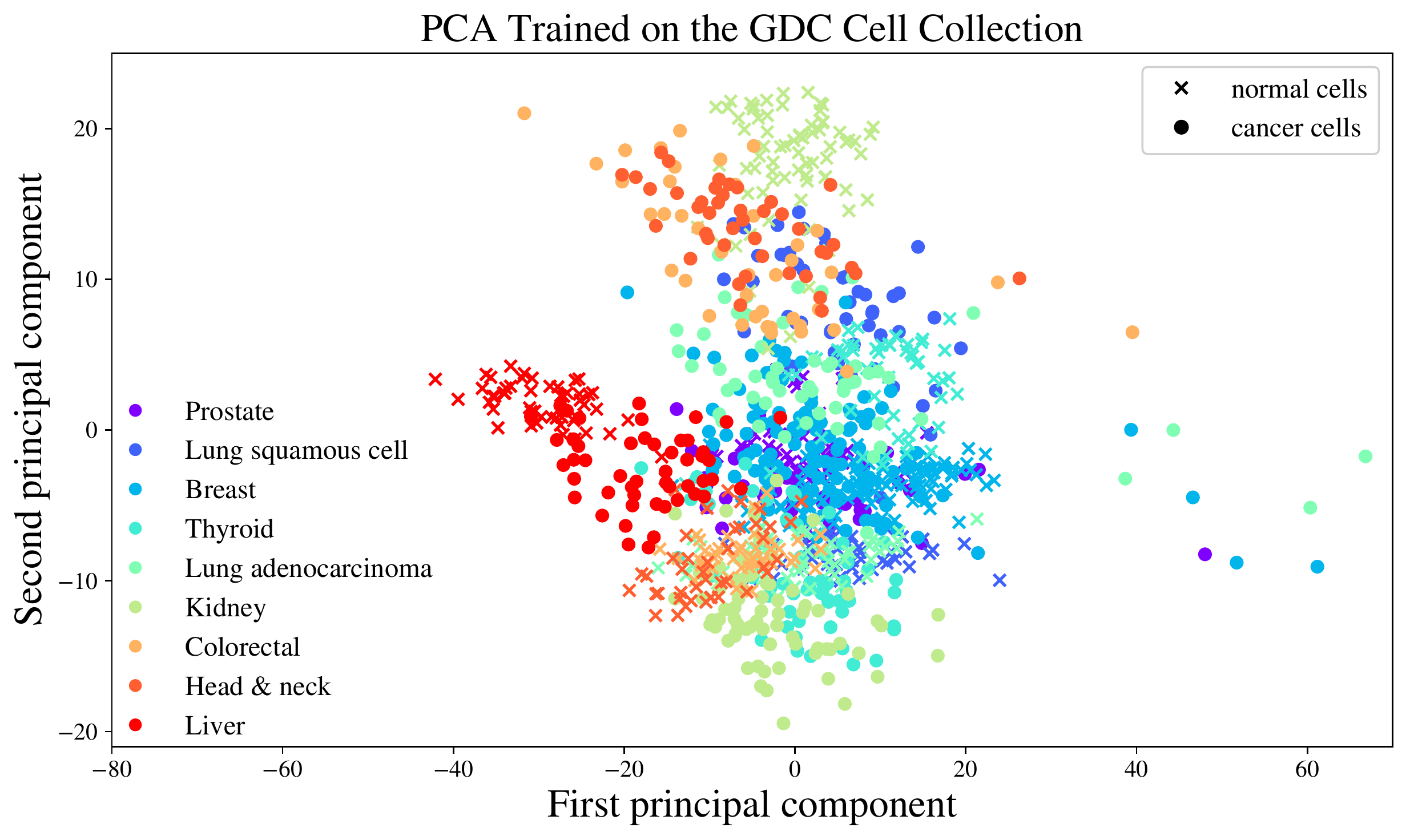}
\end{center}
\vspace{-0.2in}
\caption{\label{fig:pca} The normal and cancer cell pairs projected on the first two principal components. The PCA was trained on the GDC cancer cell collection. }
\end{figure}

%
%----------------------------------------------------
The autoencoder is a nonlinear generalization of principal components analysis (PCA) \cite{Hinton2006}. PCA finds the directions of largest variances in the dataset and projects each data point on each of these directions through linear transform. In an autoencoder if a linear transfer function is used as an activation function, a single layer autoencoder is similar to PCA.  %JEA I don't understand the previous sentence. Are you trying to say that the nonlinear activation function is needed in a single layer network to employ a nonlinear transformation?
Using the same training scheme as the autoencoder, we first applied PCA on the GDC tumor collection. Then the top two eigenvectors of the PCA are used to project the normal and cancer cell pairs on the first two principal components shown in Figure~\ref{fig:pca}. We can see that  clusters of normal and cancer cells of different tumor types are formed and overlapped on the first two principal components. A close look on the second principal component shows well separated normal cells from their paired cancer cells of three tumor types: kidney, colorectal and head \& neck. The linear projection of PCA using the directions of the largest variance (i.e. the first principal component) does not necessarily lead to distinguishing clusters of desired tumor types. PCA is often combined with other machine learning methods to demonstrate its strength of dimension reduction \cite{Huang2012}\cite{Shen2006}.

%

%The autoencoder is a nonlinear generalization of principal components analysis (PCA) \cite{Hinton2006}. PCA finds the directions of largest variance in the dataset and projects each data point on each of these directions through linear transform. These directions are eigenvectors of a dataset, generated through eigen decomposition. Each eigenvector has a corresponding eigenvalue that indicates the relative amount of variance in the data projected along its direction. 

%In the autoencoder, data are projected onto hidden nodes in the latent space, where each hidden node represents a projection. Similar to the eigenvalues, the autoencoder node saliency (ANS) \cite{Fan2019} ranks the hidden projections according to their ability to classify a given binomial property.

%----------------------------------------------------
%
\subsection{Identifying best classifying nodes}
%
%----------------------------------------------------
After training the autoencoder, we acquire optimal weight $\bm{W}^*$ and bias $\bm{b}^*$ for the GDC tumor collection. We can then generate latent representations for any sample points in the collection of pairs of normal and cancer cells. There are 1000 hidden nodes in the autoencoder. For each hidden node $s$ for $s=1,\dots, 1000$, we transform the cell data of GDC features to their latent representations using the activation function (\ref{eq:activation}). Each best node $s$ corresponds to the weight vector $\bm{w}^*_s$ that connects the input and the hidden layer, and determines how each gene in the input layer influenced the activation values of the node. Therefore, we are interested in identifying the hidden nodes that best classify normal and cancer cells for different tumor types. We first apply supervised node saliency (SNS) to the latent representations of the data subset intended for each learning task. For example, to differentiate the breast cancer cells from the normal cells, we collect the 112 pairs of tumor samples of the breast normal and breast cancer and construct their latent representations $\bm{A}$. That is $\bm{A} = [\bm{a}_s]_{s=1,\dots, m}$. Then using (\ref{eq:sns}), (\ref{eq:NED}) and (\ref{eq:NEDclass}), respectively, we compute the autoencoder node saliency values: SNS that orders the hidden nodes; NED, $\text{NED}_0$ and $\text{NED}_1$ that interpret the classification distribution. 

\begin{figure}[th]
\begin{center}
\includegraphics[width = \columnwidth, trim=0.45cm 0.3cm 0.45cm 0.3cm, clip=true]{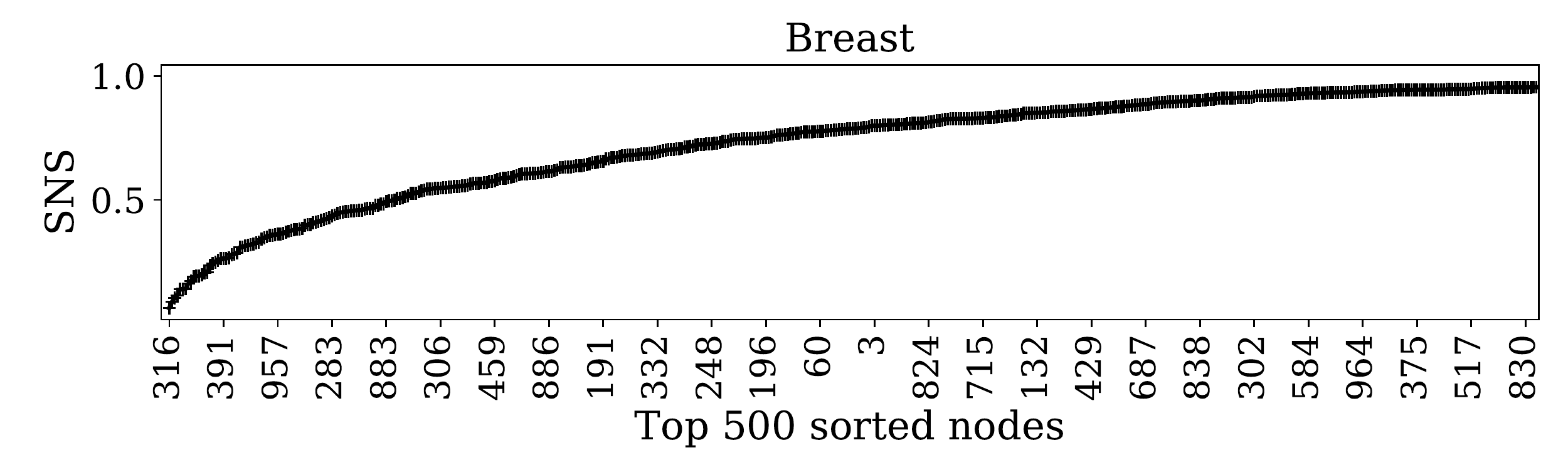}
\end{center}
\vspace{-0.2in}
\caption{\label{fig:SNSc32} Top 500 hidden nodes ranked by the supervised node saliency (SNS) on the breast dataset. The original node numbers appear every 20 nodes. The best node for classifying the normal and cancer cells of breast is Node 316. }
\end{figure}

\begin{figure}[th]
\begin{center}
\begin{tabular}{cc}
\includegraphics[width = 0.4\columnwidth, trim=0.3cm 1.9cm 0.45cm 0.3cm, clip=true]{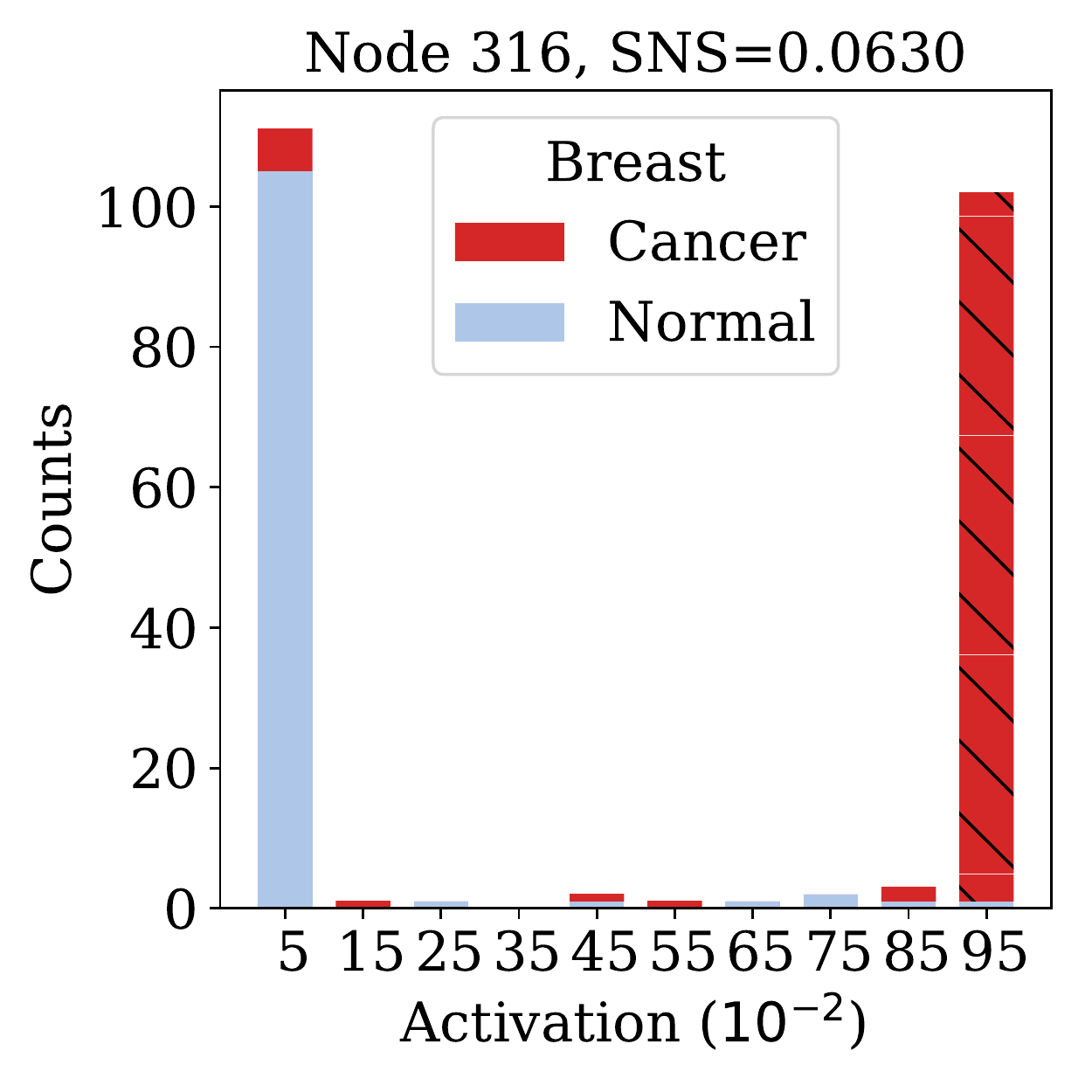}
&\includegraphics[width = 0.4\columnwidth, trim=0.3cm 1.9cm 0.3cm 0.3cm, clip=true]{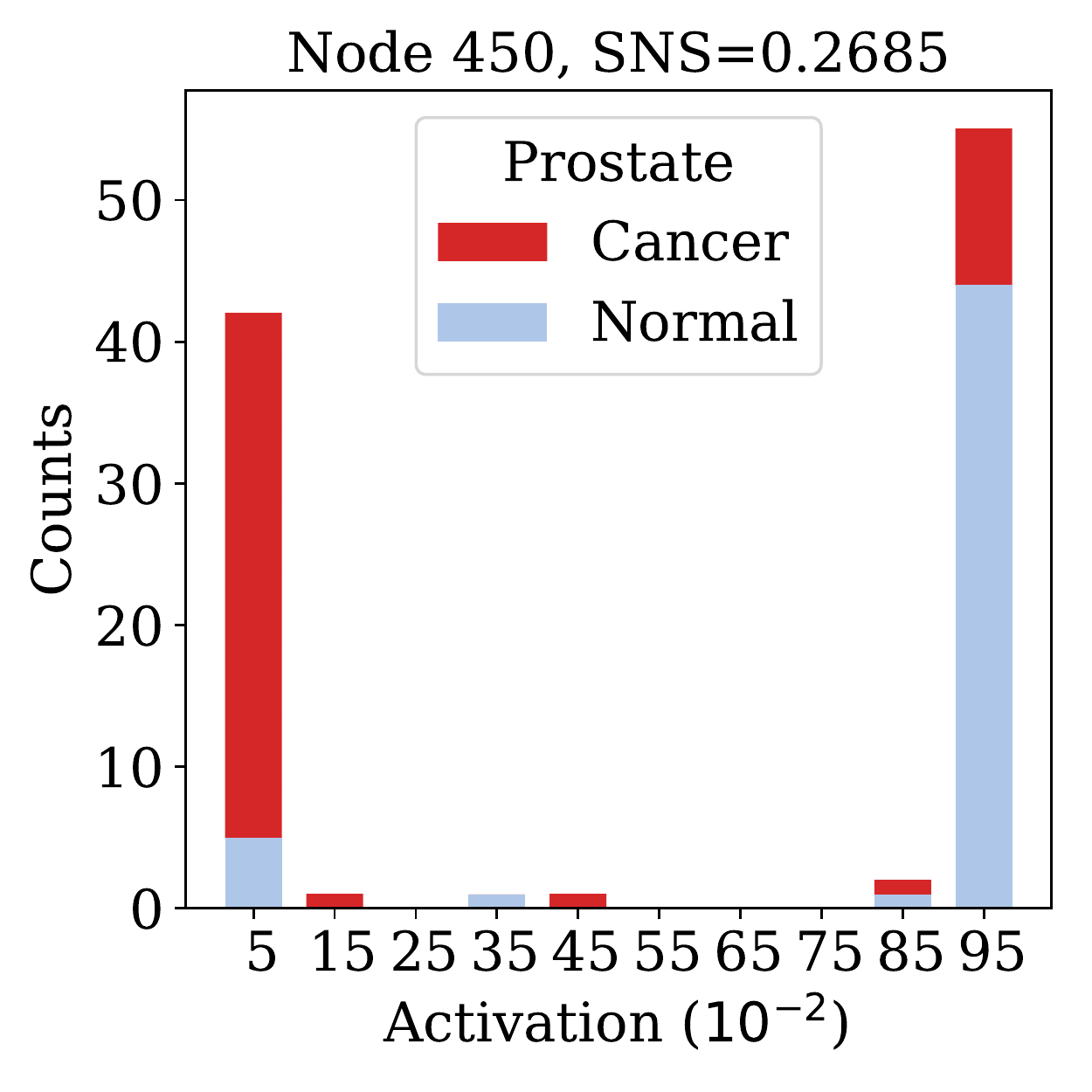}\\
\includegraphics[width = 0.4\columnwidth, trim=0.3cm 1.9cm 0.45cm 0.3cm, clip=true]{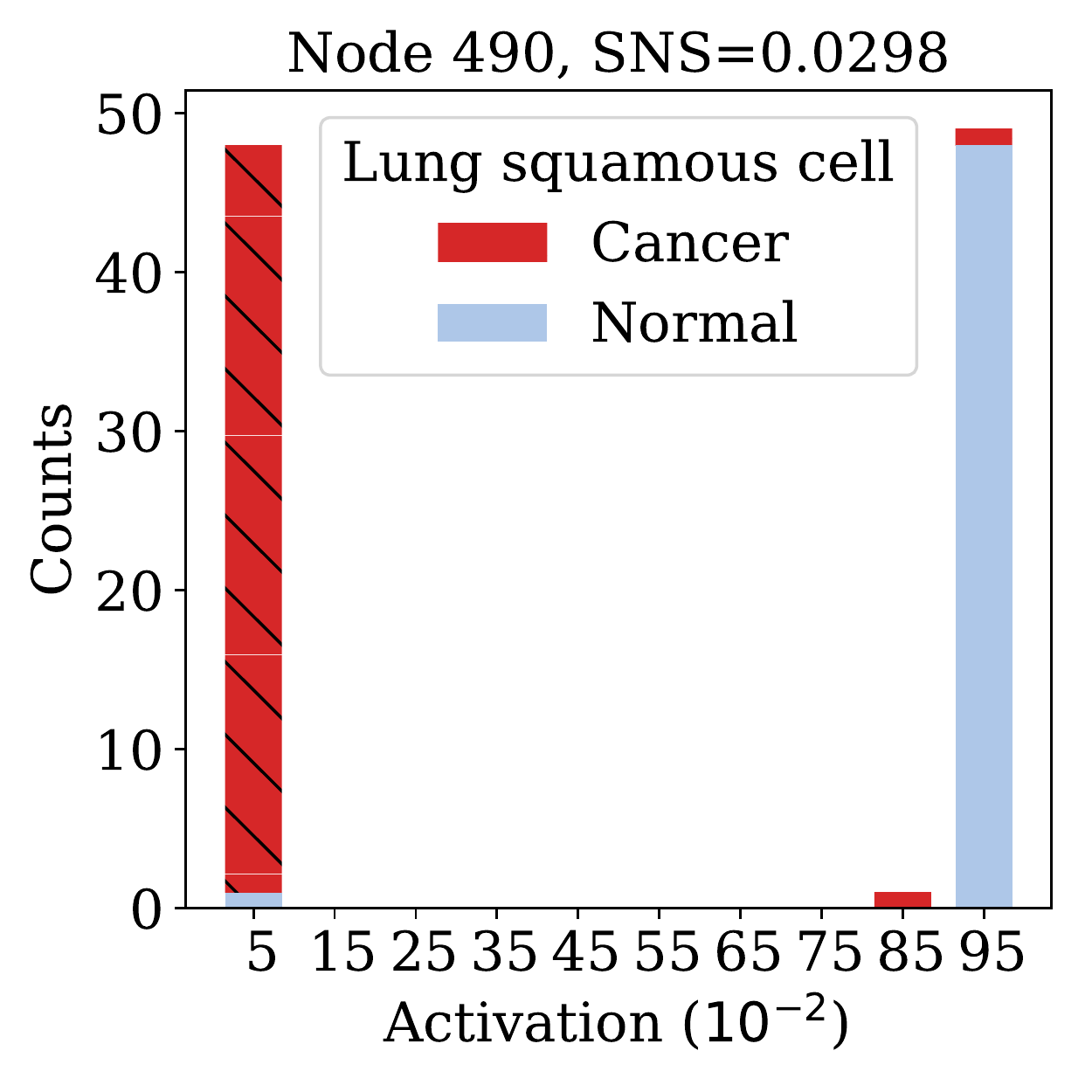}
&\includegraphics[width = 0.4\columnwidth, trim=0.3cm 1.9cm 0.3cm 0.3cm, clip=true]{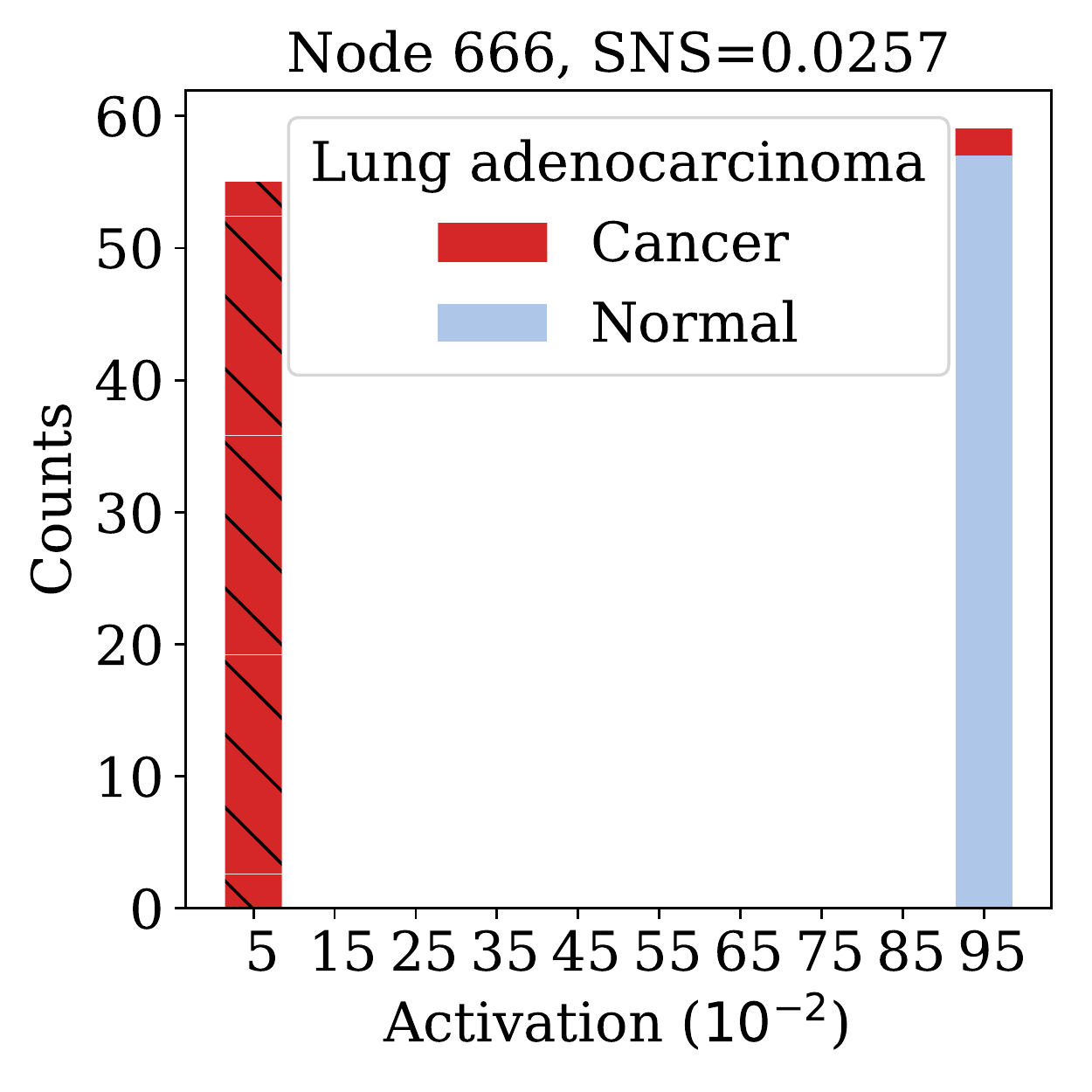}\\
\includegraphics[width = 0.4\columnwidth, trim=0.3cm 1.9cm 0.45cm 0.3cm, clip=true]{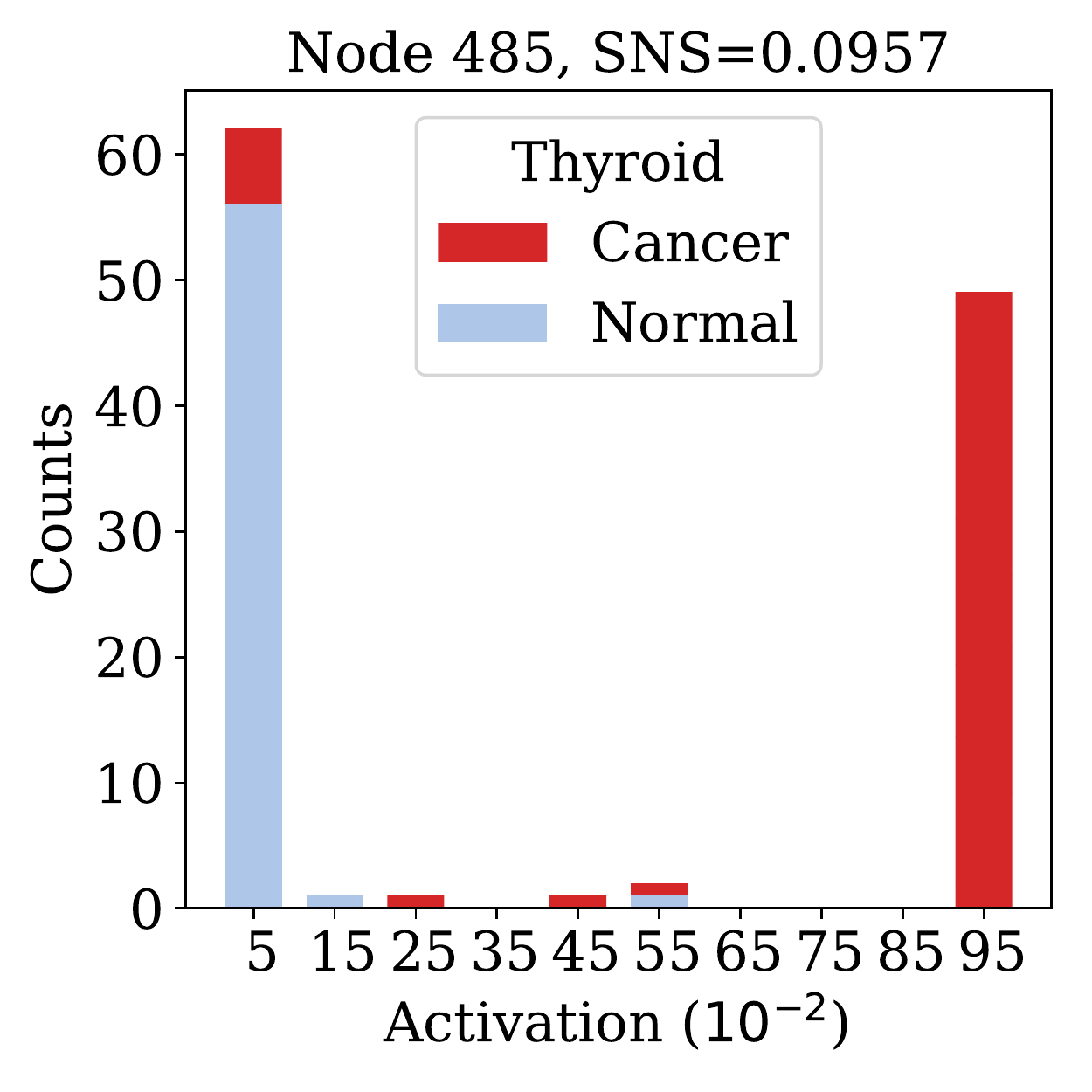}
&\includegraphics[width = 0.4\columnwidth, trim=0.3cm 1.9cm 0.3cm 0.3cm, clip=true]{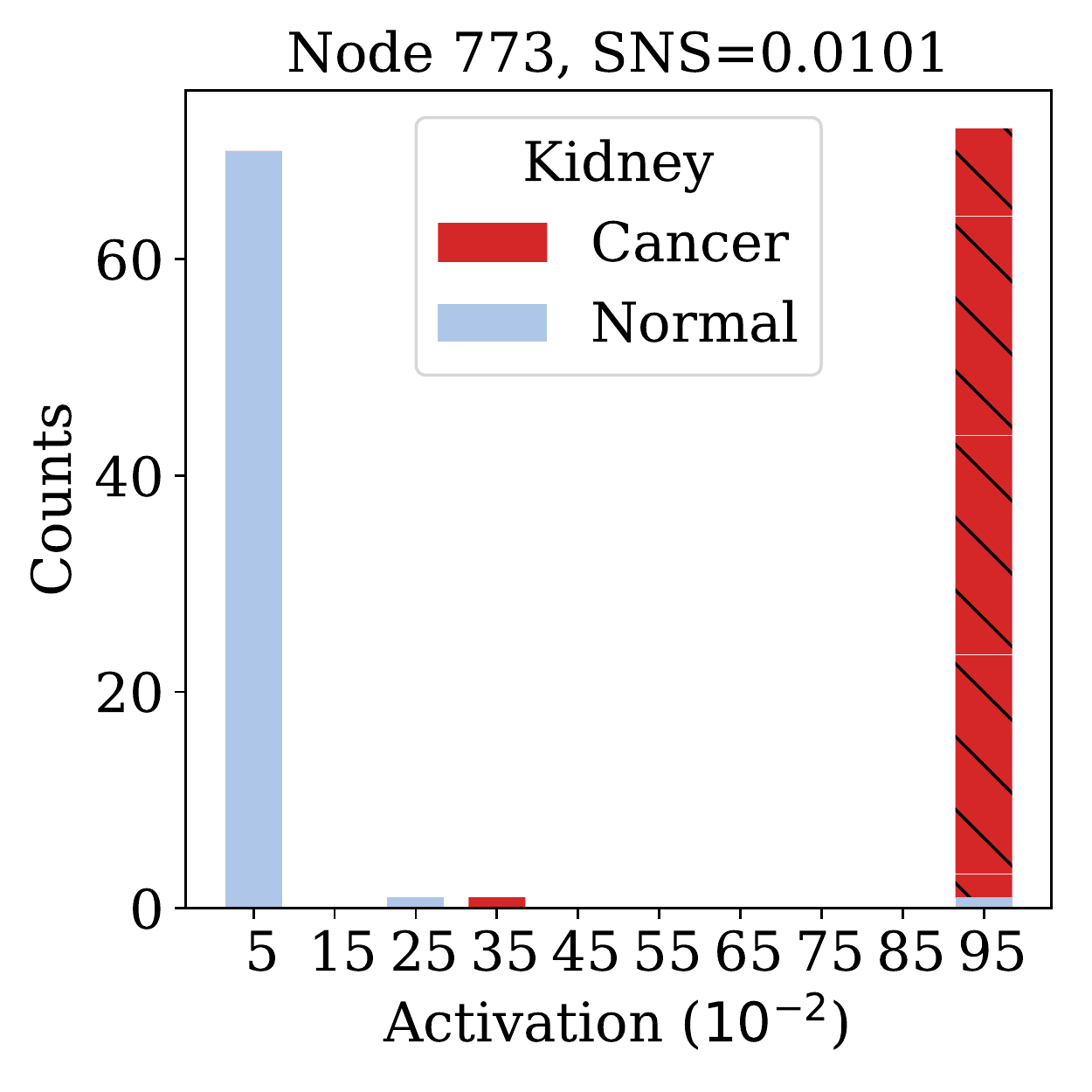}\\
\includegraphics[width = 0.4\columnwidth, trim=0.3cm 1.9cm 0.45cm 0.3cm, clip=true]{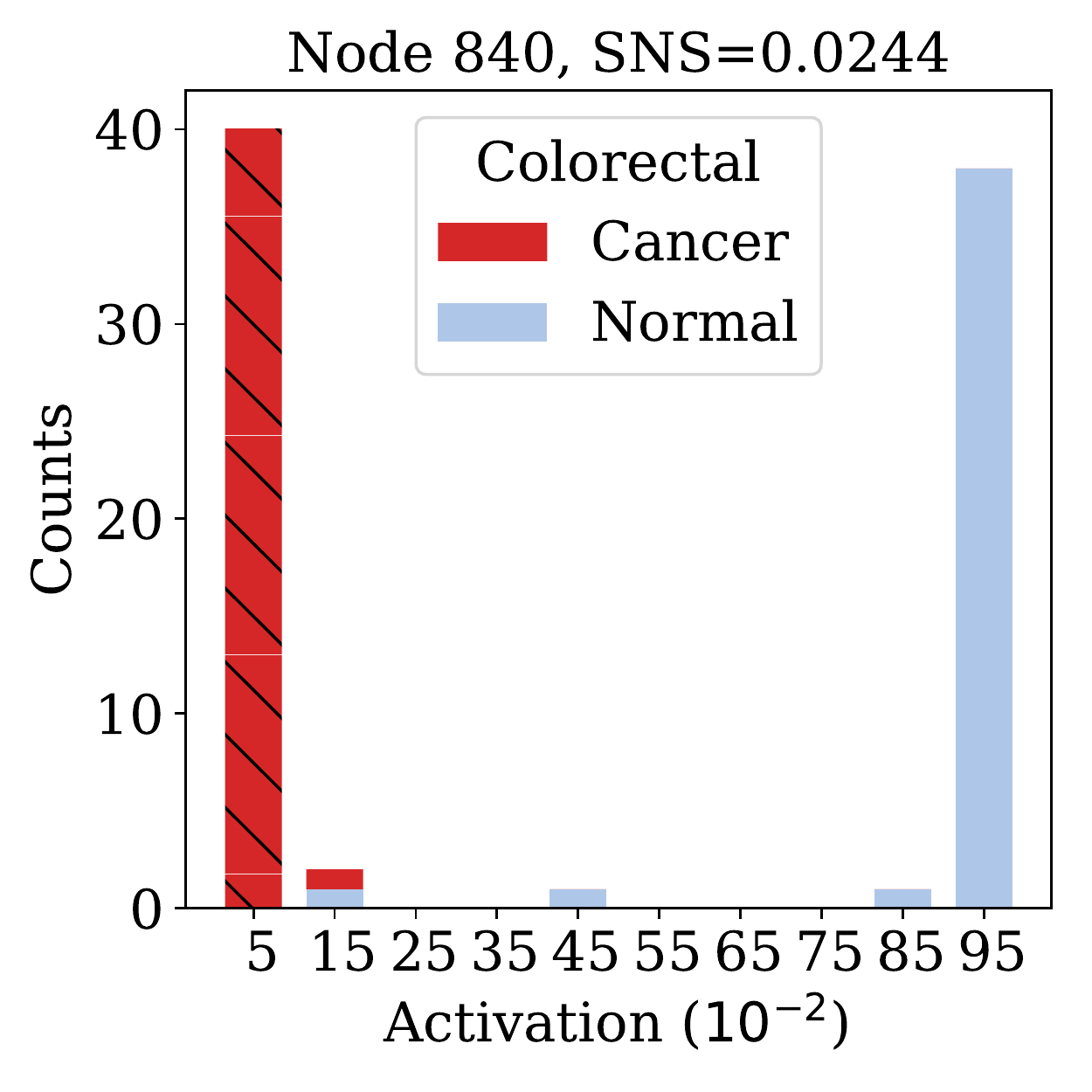}
&\includegraphics[width = 0.4\columnwidth, trim=0.3cm 1.9cm 0.3cm 0.3cm, clip=true]{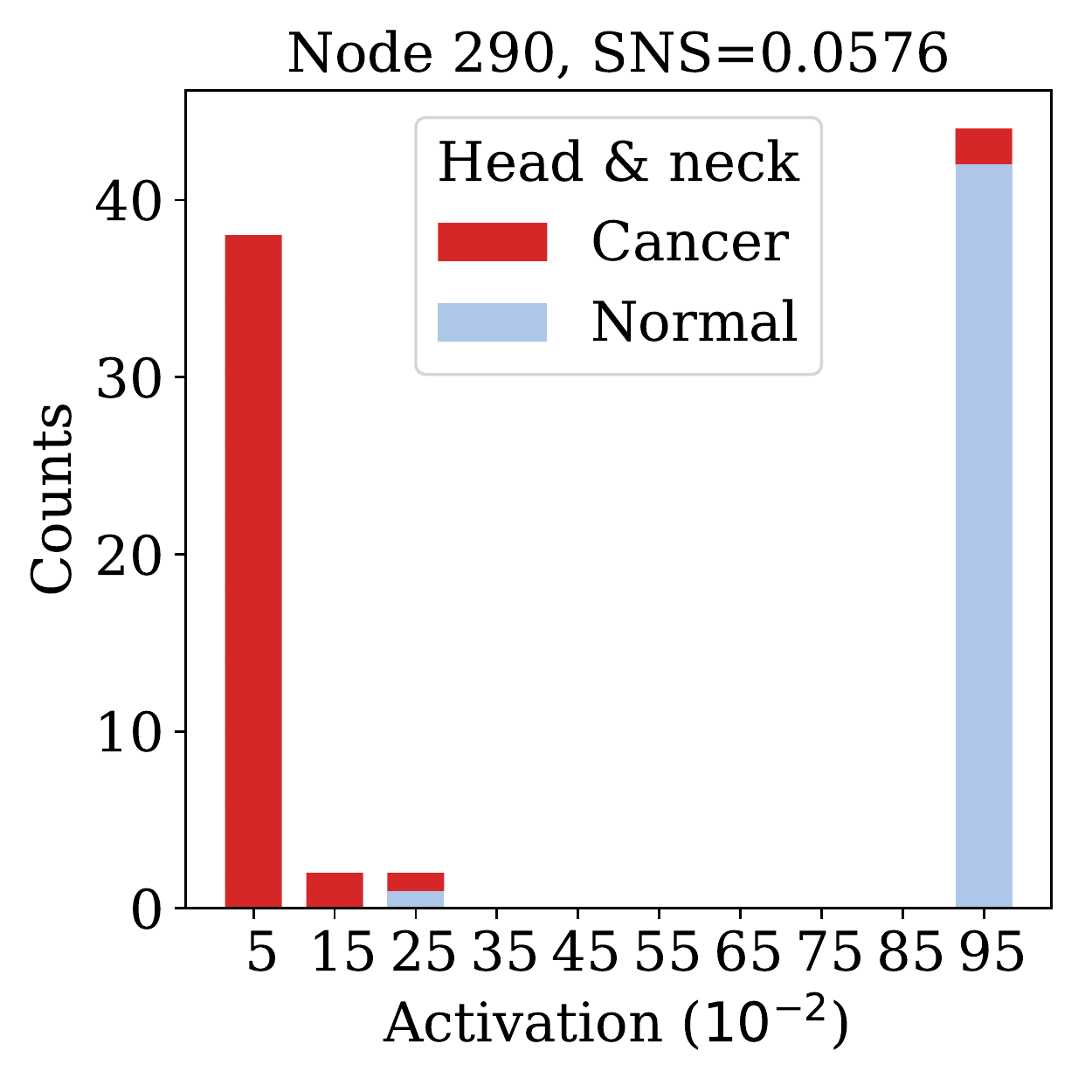}\\
\includegraphics[width = 0.4\columnwidth, trim=0.3cm 0.5cm 0.45cm 0.3cm, clip=true]{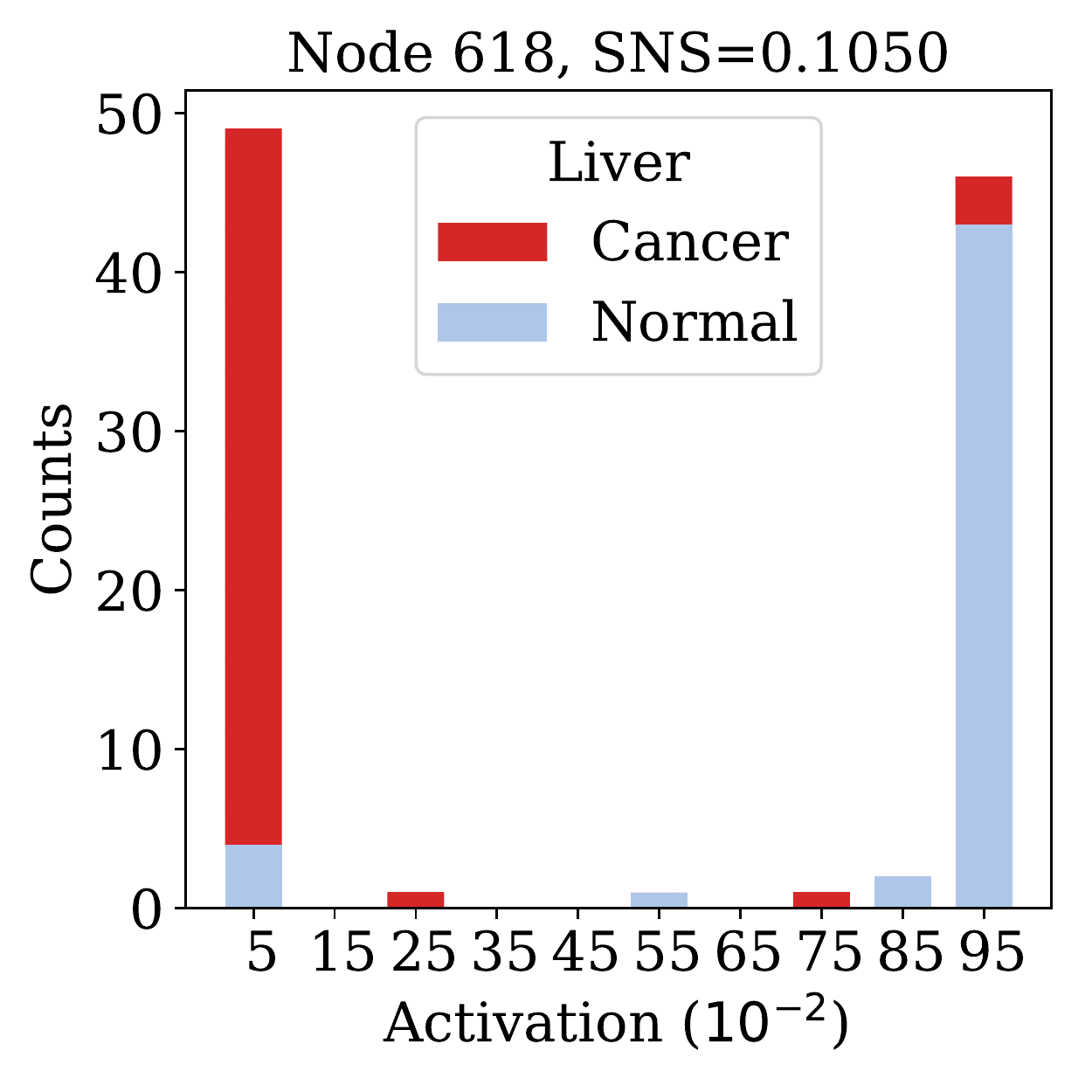}
& \includegraphics[width = 0.4\columnwidth, trim=0.3cm 0.5cm 0.3cm 0.3cm, clip=true]{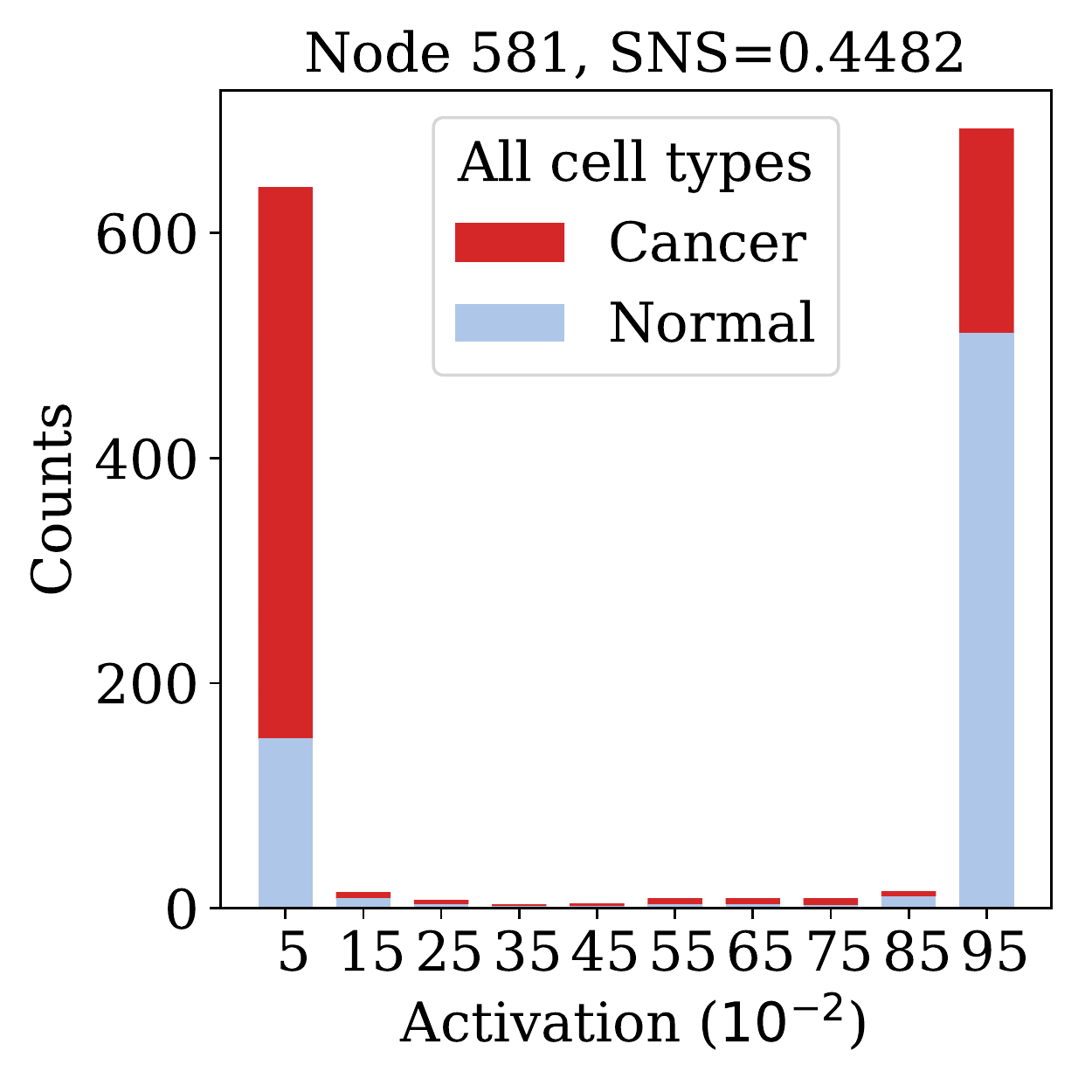}
\end{tabular}
\end{center}
\vspace{-0.17in}
\caption{\label{fig:bestNodes} Best classifying nodes of the trained autoencoder for distinguishing between normal and cancer cells in each tumor type.}
\end{figure}

Figure~\ref{fig:SNSc32} displays the top 500 (out of 1000) hidden nodes ranked in increasing order of their SNS values on the normal and cancer cells of breast. The SNS curves for most binary classification tasks in the dataset are similar to the curve in the figure. Only the top few nodes have low SNS values close to zero, while the majority of the rest nodes are close to 0.8. The best node for classifying the normal and cancer cells of breast is node 316. The histogram of node 316 that describes the distribution of the normal and cancer cells of breast is shown in the first histogram in Figure~\ref{fig:bestNodes}. We observe that the normal and cancer cells of breast are very well separated in node 316 with a small percentage of cancer cells allocated at the cluster of the normal cells. This result is consistent with the cell clusters shown in the t-SNE plot (Figure~\ref{fg:tSNE}). The t-SNE indicates that there exist a small cluster of mixed cancer cells next to the normal breast cells. There are also a few breast cancer cells that are similar to normal cells. These cancer cells are difficult to differentiate.

\begin{figure}[thp]
\begin{center}
\includegraphics[width = \columnwidth, trim=0.2cm 0.3cm 0.45cm 0.3cm, clip=true]{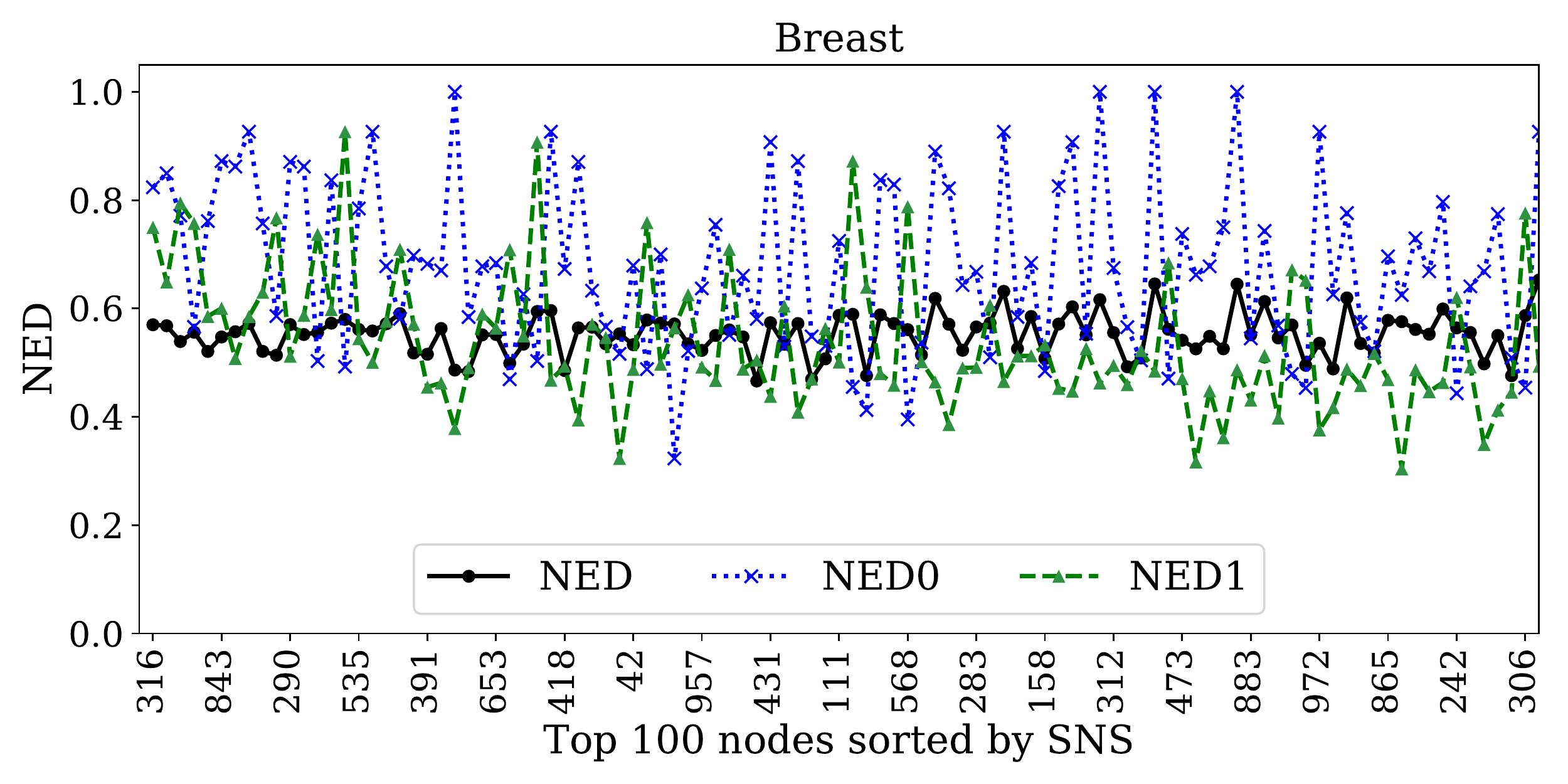}
\end{center}
\vspace{-0.2in}
\caption{\label{fig:NEDc32} NED values of the top 100 hidden nodes ranked by the supervised node saliency (SNS) on the breast dataset. The original node numbers appear every five nodes. The NED values on the top six nodes satisfy both NED $<\text{NED}_0$ and NED $<\text{NED}_1$, the property of good classifying nodes.}
\end{figure}

The next step of the autoencoder node saliency is to verify the property of a good classifying node on the top SNS nodes. As discussed in Section~\ref{sec:NED}, a good classifying node meets both conditions: NED $<\text{NED}_0$ and NED $<\text{NED}_1$. Figure~\ref{fig:NEDc32} displays the NED values of the top 100 hidden nodes ranked by the SNS values for the breast cell pairs. The top six nodes, including node 316, satisfy the property of a good classifying node, while the rest of the nodes do not. 

%SNS ranks the hidden nodes for a learning task in an increasing order. The best classifying nodes have the lowest SNS values. 

%

The SNS values allow us to compare different classification tasks. Since SNS is computed by comparing the activation distribution to a fixed reference distribution (\ref{eq:binary}), histograms with a better classification have lower SNS values. Histograms of the best classifying nodes that separate cancer from normal are shown in Figure~\ref{fig:bestNodes}, each for one of the nine tumor types (in Table~\ref{tb:dataDescription}). The histogram of the best classifying node (node 581) for all tumor types combined is shown in the last figure. Their corresponding SNS values are also listed on the top of the histogram. Distinguishing between kidney renal clear cell carcinoma and normal kidney (node 773) is the easiest task among the ten tasks, with the lowest SNS at 0.0101. Only a small number of normal kidney cells mixed with the cluster of the kidney renal clear cell carcinoma. Among individual tumor types, prostate adenocarcinoma is the most difficult to separate from prostate normal cells. The best classifying node for prostate is node 450 with SNS at 0.2685. This is again consistent with the t-SNE plot (Figure~\ref{fg:tSNE}) where there is barely a space between the cluster of prostate adenocarcinoma and the cluster of prostate normal. The most difficult classification is to separate cancer from normal considering all tumor types at the same time. Node 581 has the largest SNS at 0.4482 and a large number of mixed cancer and normal cells. 

\begin{figure}[th]
\begin{center}
\begin{tabular}{cc}
\includegraphics[width = 0.45\columnwidth, trim=0.2cm 1.2cm 0.2cm 0.2cm, clip=true]{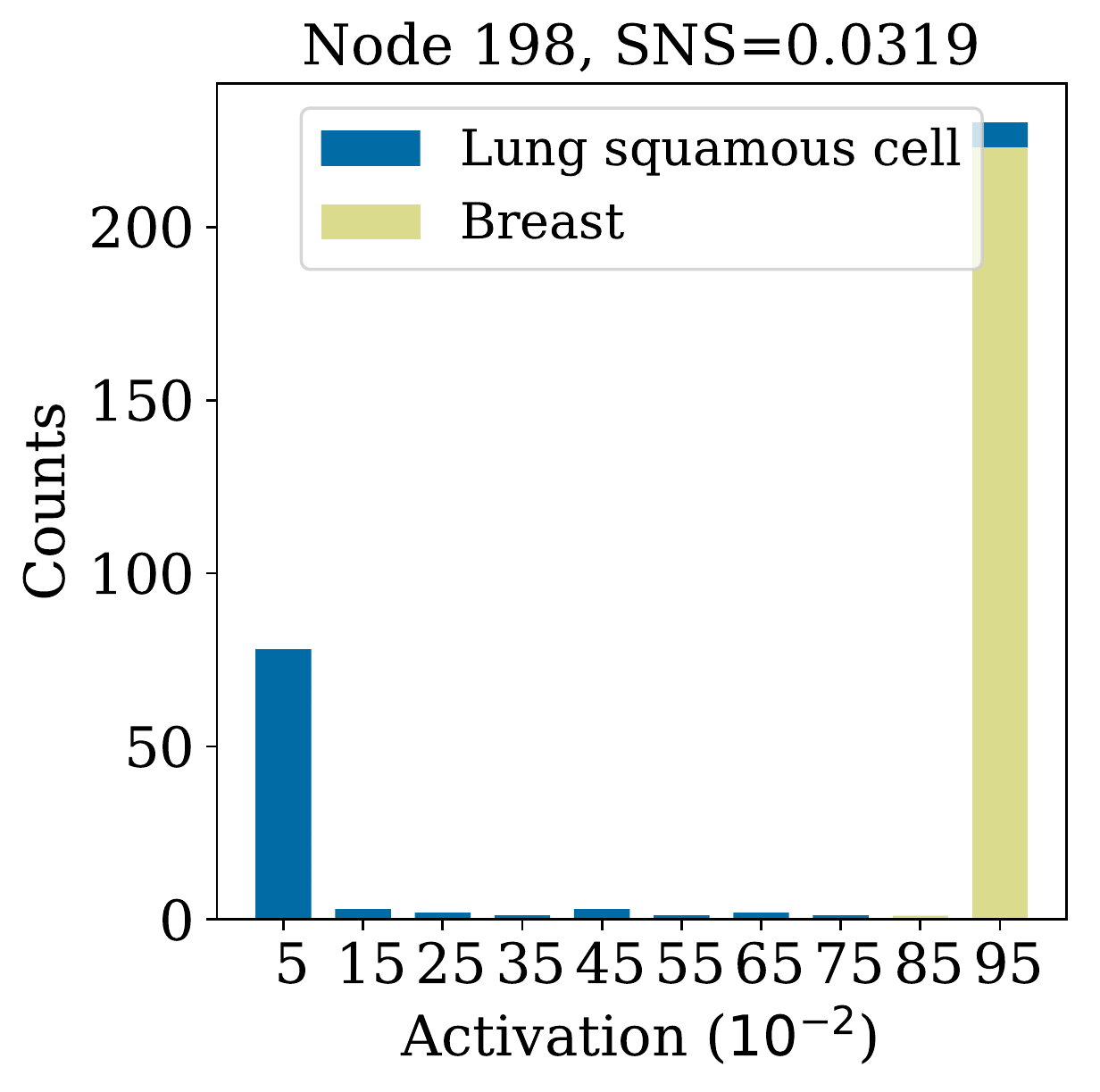}
&\includegraphics[width = 0.45\columnwidth, trim=0.2cm 1.2cm 0.2cm 0.2cm, clip=true]{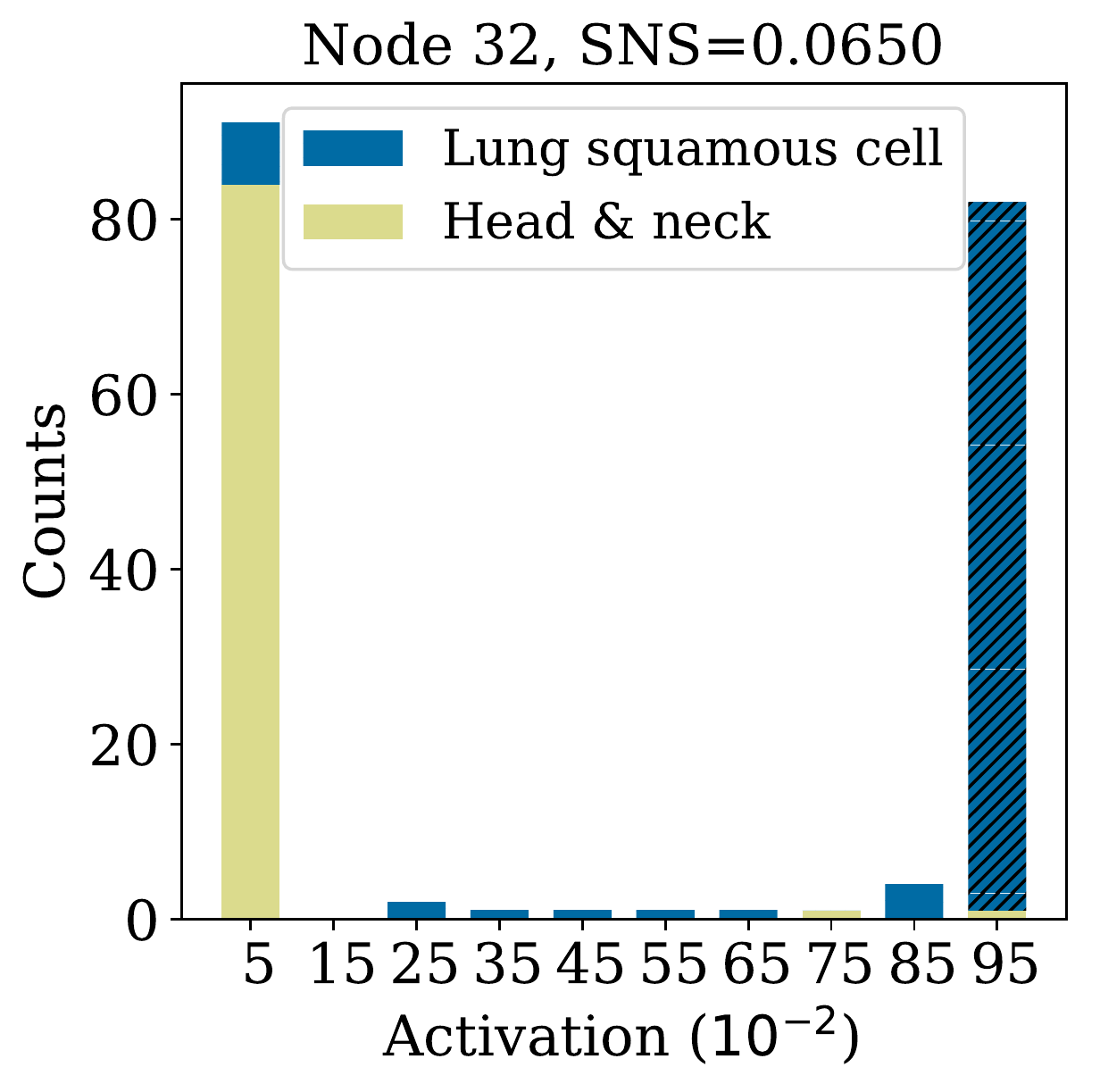}\\
%&\includegraphics[width = 0.46\columnwidth, trim=0.2cm 1.4cm 0.2cm 0.2cm, clip=true]{sns_c28c42.pdf}\\
\includegraphics[width = 0.45\columnwidth, trim=0.2cm 1.2cm 0.2cm 0.2cm, clip=true]{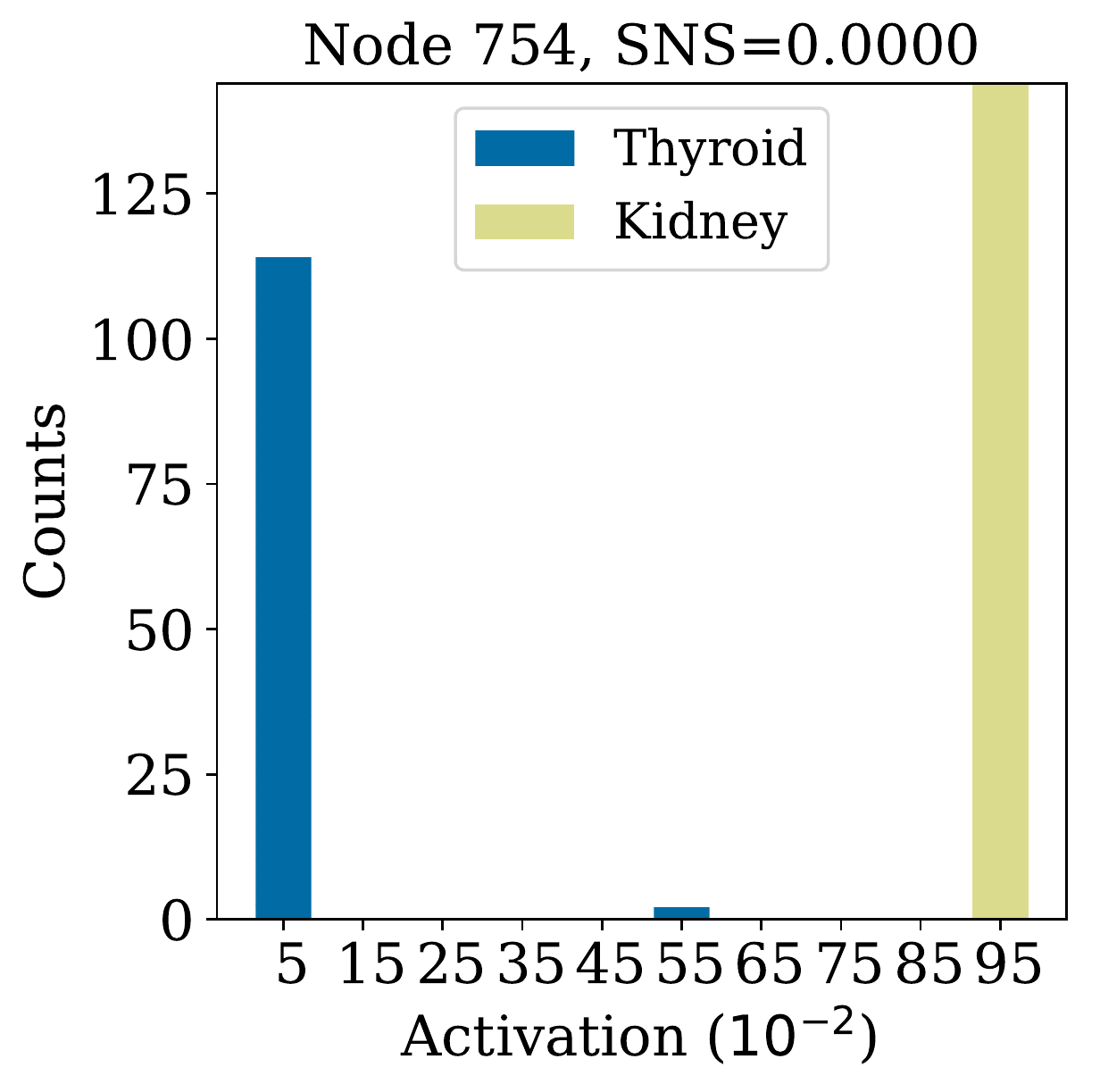}
&\includegraphics[width = 0.45\columnwidth, trim=0.2cm 1.2cm 0.2cm 0.2cm, clip=true]{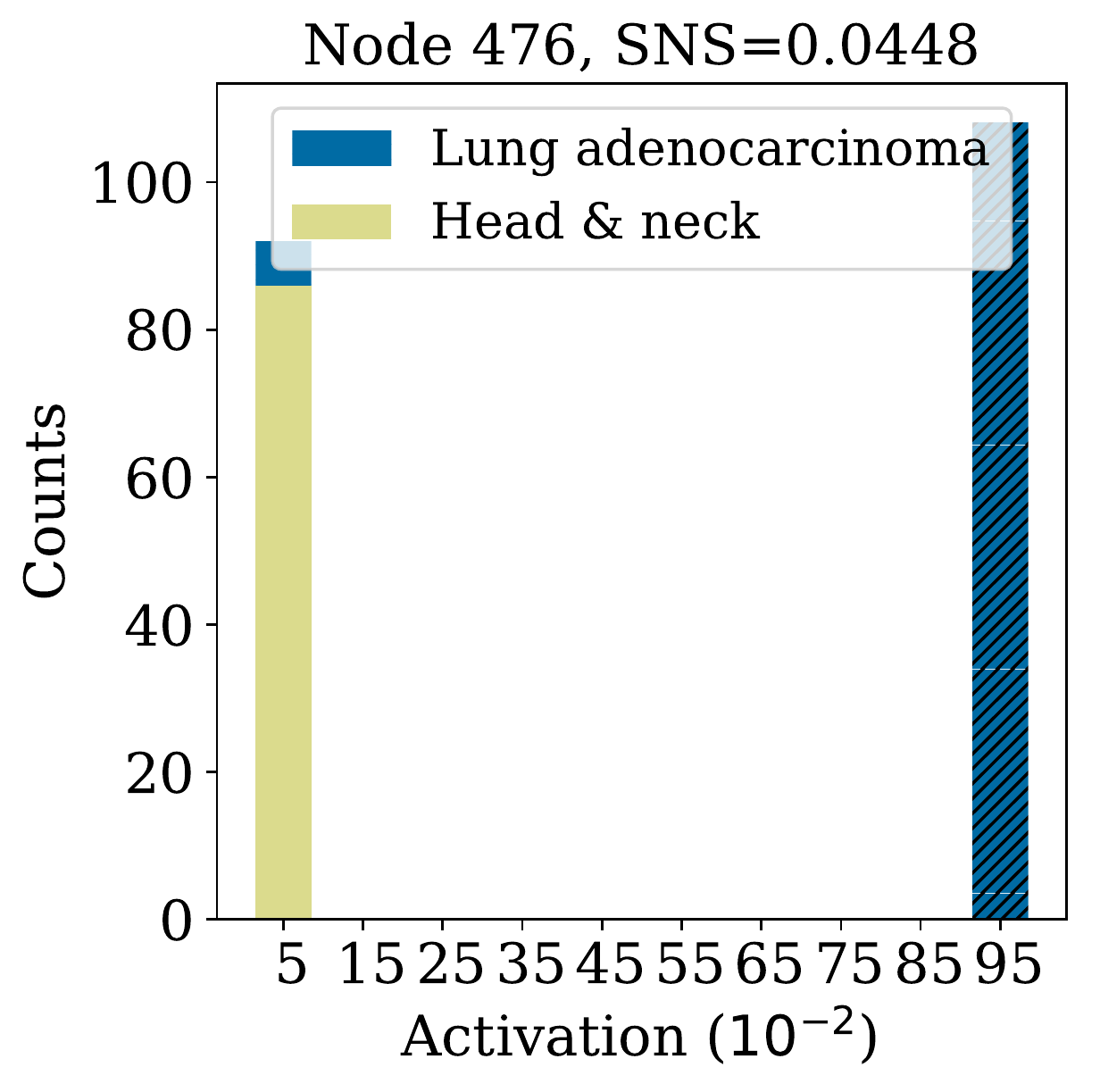}\\
\includegraphics[width = 0.45\columnwidth, trim=0.2cm 0.2cm 0.2cm 0.2cm, clip=true]{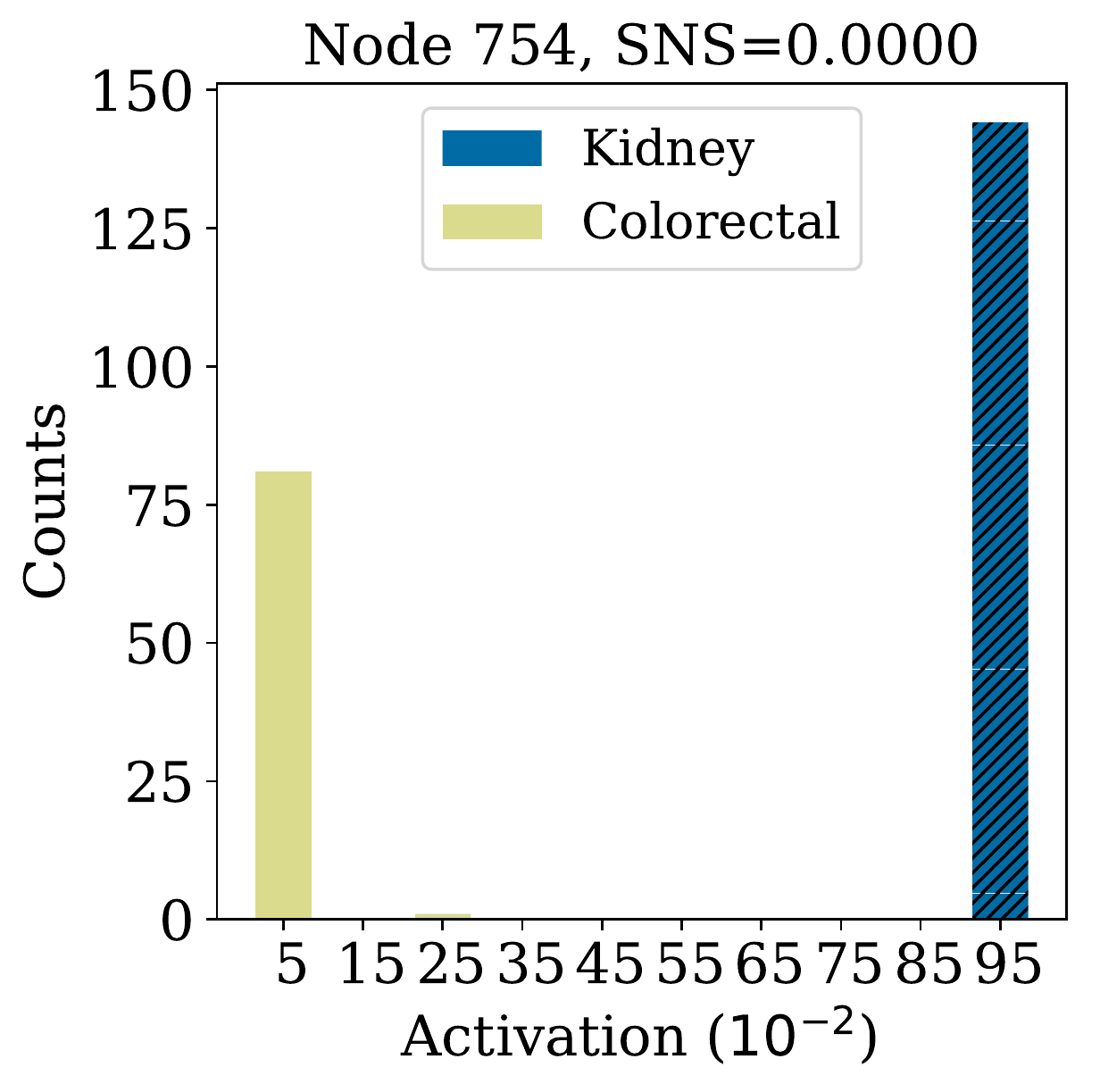}
&\includegraphics[width = 0.45\columnwidth, trim=0.2cm 0.2cm 0.2cm 0.2cm, clip=true]{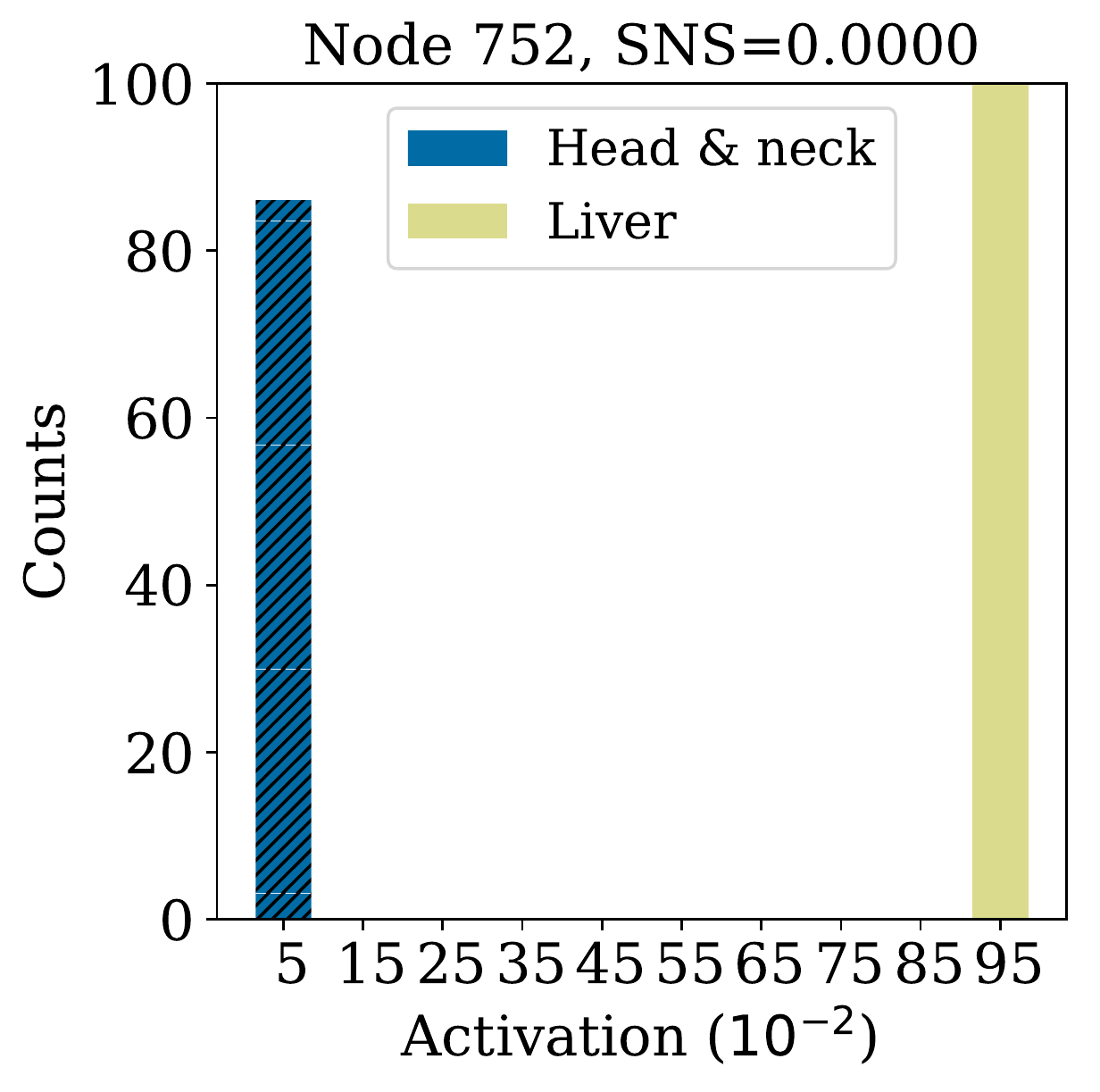}
\end{tabular}
\end{center}
\vspace{-0.15in}
\caption{\label{fig:bestNodesCell2Cell} Best classifying nodes of the trained autoencoder for distinguishing between tumor types together with their corresponding normal cells. We display six random pairs of the tumor types.}
\end{figure}

Research indicates that mutational processes vary among tumors and cancer types \cite{Martincorena2015, Vogelstein2013}. Tumor types play a significant role in understanding how certain genes are altered by mutations \cite{Vogelstein2013}. To further demonstrate the value of the autoencoder node saliency method, we randomly select six pairs of tumor types (including both the tumors and normal cells) and display their distribution on their best classifying nodes in Figure~\ref{fig:bestNodesCell2Cell}. 
%JEA is this tumor and normal? I assume so and make this explicit above. YJF it is the two combined.
Node 198 with SNS at 0.0319 is the best classifying node for separating lung squamous cell carcinoma and breast invasive carcinoma, considering both tumor and normal tissue. The smallest SNS among these tumor type pairs is node 754 and 752, having SNS at its minimum, with the least misclassified cells in the latent representation of the hidden node. The node 754 is the best classifying node for differentiating kidney from both thyroid and colorectal tumors; while the node 752 is the best classifying node for separating liver from head \& neck. These tumor types are far away from each other in their t-SNE plot. 

%JEA suggest removing unless there is more information on this: Lungs and breasts are physically close to each other. Their cancer cells could possess higher similarity. 

\begin{figure}[ht]
\begin{center}
\includegraphics[width = 0.45\columnwidth, trim=0.2cm 0.2cm 0.2cm 0.2cm, clip=true]{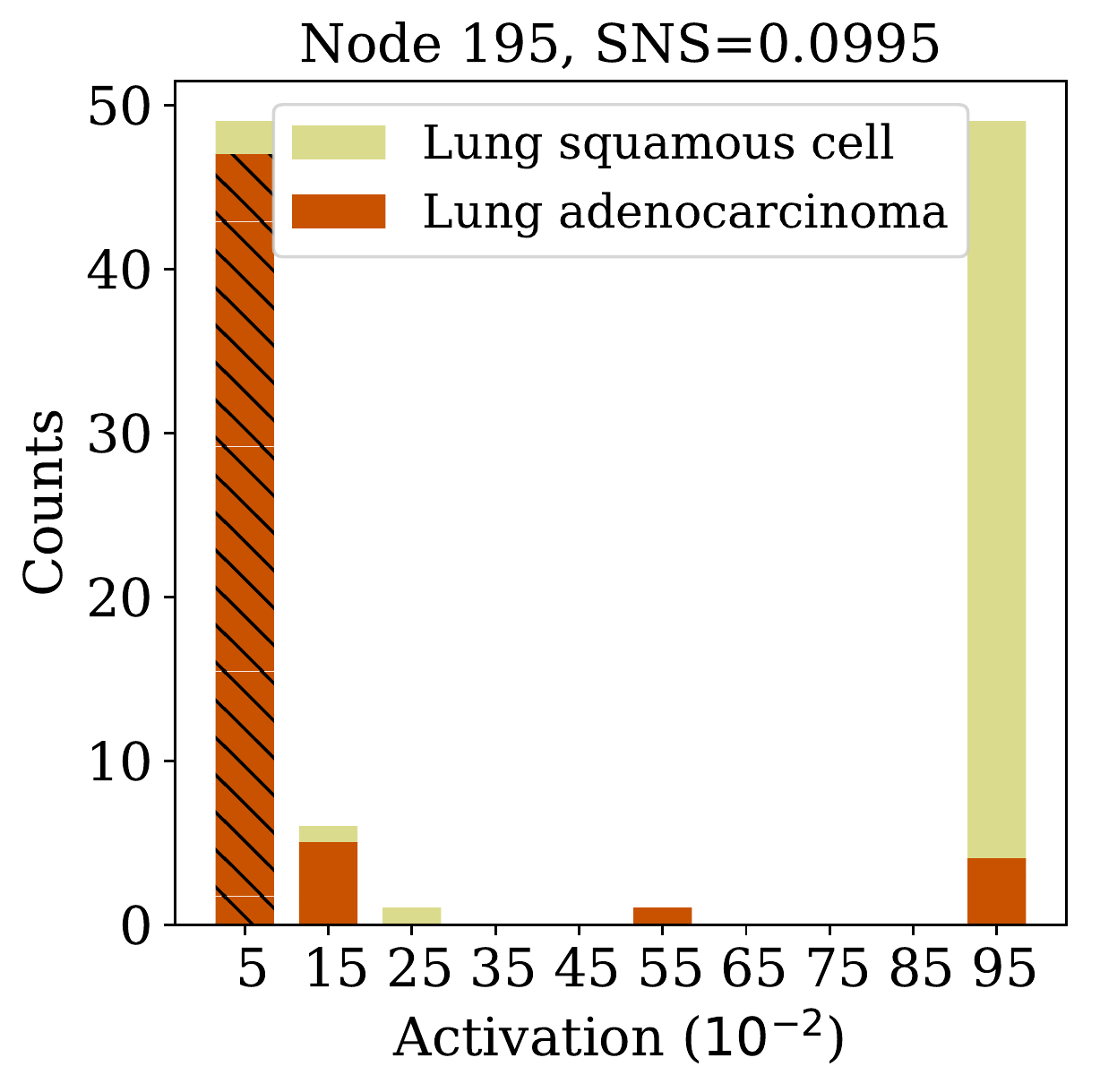}
\end{center}
\vspace{-0.25in}
\caption{\label{fig:snsLung} The best classifying node that separates the two lung cancers, Lung Squamous Cell Carcinoma and Lung Adenocarcinoma. Their corresponding normal cells are ignored.}
\end{figure}

The normal lung cells could turn into lung squamous cell carcinoma or lung adenocarcinoma. Since their normal cells are from the same category, we only consider the latent representations from the two lung cancer types for computing SNS. The histogram of the best node is shown in Figure~\ref{fig:snsLung}. We find that node 195 best classifies the two lung cancers with SNS at 0.0995. 
\begin{figure}[htp]
\begin{center}
\includegraphics[width = 0.9\columnwidth, trim=0.2cm 0.2cm 0.2cm 0.2cm, clip=true]{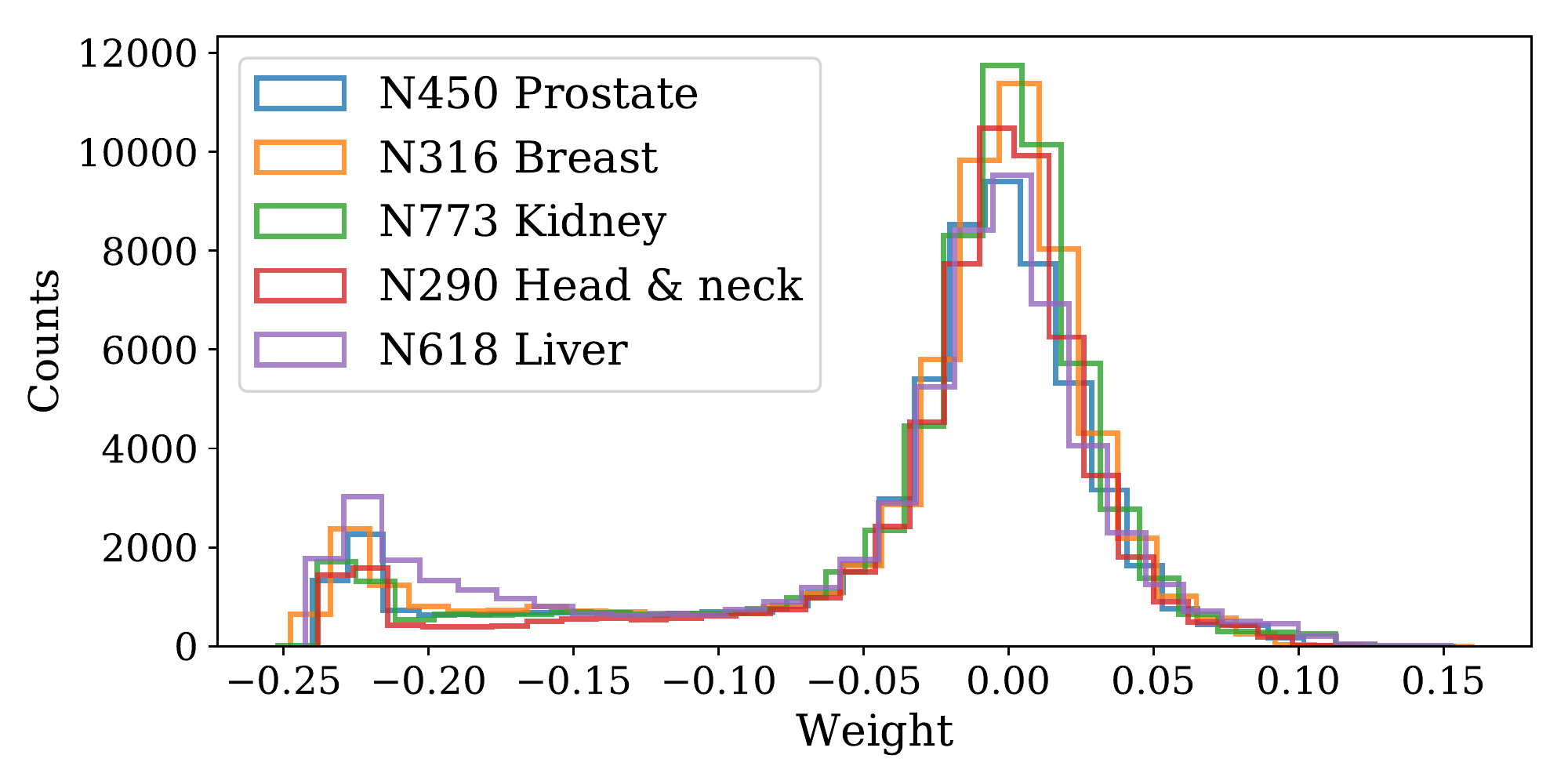}
\end{center}
\vspace{-0.25in}
\caption{\label{fig:w} The distributions of weights for all 60,483 transcripts to the best classifying nodes. Each node separates cancer cells from normal cells of a different tumor type. We randomly select five tumor types for the plot. }
\end{figure}
%
%we still lack quantitative information about the role of different factors in the evolution of normal cells into cancer cells

Recall that PCA is able to separate cancer from normal for three cancer types (Figure~\ref{fig:pca}) at the second principal components. The eigenvalues of PCA indicates the data variances along the principal components. However, whether other principal components contain clusters of different tumor types is still to be inspected. The autoencoder presents a useful alternative to PCA, which captures the t-SNE similarity. Moreover, through the use of the autoencoder node saliency method we can efficiently extract interesting features that could describe the visualized patterns.
 
%We still need other methods to determine whether PCA is able to provide any clustering of different tumor types at other top principal components. %Although we may find the shared factors separating cancers from normals, it is not straight forward to explore additional principal components to find a direction that separates within different tumor types without using other learning methods. 

The identified autoencoder nodes not only reflect the high similarity of the tumor types in the t-SNE plot, but also point out distinctive genes without using any class labels, which t-SNE is not able to reveal. The distribution of weights for all 60,483 transcripts to a single node approximately resembled a bi-model distribution with one mode centered around zero and a second centered around -0.225 as shown in Figure~\ref{fig:w}. We randomly select five tumor types for display. The majority of the transcripts have zero or low weights in a hidden node; while a small portion of the genes have high negative weights. Each node has a unique distribution of weights. We can withdraw common genes among different tumor types and extract speciality genes from individual tumor types for further analysis. The weights of the best classifying nodes provide links from the latent representations back to the original genes that stimulate the evolution of normal cells into cancer cells.

%===========================================
%
\section{Conclusion}
%
%===========================================

An autoencoder model is trained on a large collection of tumor samples (11,574) represented by 60,483 measured transcripts from the GDC data portal. It provides a generalized and unsupervised latent representation of cancer cells. We leverage the scalable learning toolkit, LBANN, to take advantage of available HPC systems, thus reducing computation time. Experimental results from LBANN show strong scaling on CPU and GPU clusters. This scalability provides opportunities for quick massive hyperparameter exploration. There are 1000 hidden nodes in the autoencoder. After training the autoencoder, each hidden node has a corresponding optimal weight and a bias. We use them to compute the latent representations of a small but non-overlapping dataset, containing 533 pairs of normal and cancer cells of multiple tumor types.  %JEA make sure we clairfy at the top what is meant by "independent" dataset.

Autoencoder node saliency (ANS) determines the best classifying node using two measurements: SNS to rank the hidden nodes and NED to verify classifying properties. The results show that the autoencoder constructed from the large collection of tumor samples is able to identify features in the small dataset of the paired normal and cancer cells. The pairs of the normal and cancer cells of the nine various cancer types are visualized using t-SNE, which reveals the similarity and dissimilarity of these data points. However, the actual distances in t-SNE are not preserved. We demonstrate that the identified best classifying nodes of the autoencoder are able to fill the missing connection between the visualized data clusters and their underlying distinctive factors. ANS found the best classifying nodes that distinguish between normal and cancer cells for each of the tumor types. It also revealed the hidden nodes that separate different tumor types. %Comparing to PCA, which is restricted to a linear transform, we demonstrate that the autoencoder is able to capture nonlinear relationships among transcripts.  %JEA not sure I fully get where this happened yet...so an explicit example here might be helpful.
%It is not straight forward for PCA to identify, which principal components correspond to genes for the different cancer types. 
Moreover, the weights of the best classifying nodes provide links from the latent representations back to the original genes. We are able to extract the speciality genes that stimulate the evolution of normal cells into cancer cells.

\section*{Acknowledgment}

LLNL-CONF-750517. We thank scientists from the Computing, Environment and Life Sciences Directorate at Argonne National Laboratory:
Maulik Shukla provided access to the GDC dataset that was generated as part of the JDACS4C project; Rick L. Stevens and Fangfang Xia gave inspiration on t-SNE visualization on the data. This work was performed under the auspices of the U.S. Department of Energy by Lawrence Livermore National Laboratory under Contract DE-AC52-07NA27344.

\bibliographystyle{IEEEtran}
\bibliography{myRef}  
\pagebreak

\end{document}